\documentclass[lettersize,journal]{IEEEtran}
\usepackage{amsmath,amsfonts}
\usepackage{algorithmic}
\usepackage{algorithm}
\usepackage{array}
\usepackage{textcomp}
\usepackage{stfloats}
\usepackage{booktabs} 
\usepackage{makecell} %
\usepackage{url}
\usepackage{verbatim}
\usepackage{multirow}
\newcommand{\first}[1]{{\color{red}\bf{#1}}}

\usepackage{graphicx}
\usepackage{subcaption}
\usepackage{amsmath}
\usepackage[colorlinks=true]{hyperref}
\usepackage{cite}

\usepackage{fontawesome5}

\begin{document}

\title{Restoration-Oriented Video Frame Interpolation with Region-Distinguishable Priors from SAM}

\author{Yan Han, Xiaogang Xu, Yingqi Lin, Jiafei Wu, Zhe Liu, Ming-Hsuan Yang~\IEEEmembership{IEEE Fellow}

 \IEEEcompsocitemizethanks{
 
 \IEEEcompsocthanksitem  Yan Han, Yingqi Lin, Jiafei Wu and Zhe Liu are with Zhejiang Lab. 
Email:\{ hanyan, linyq, wujiafei, zhe.liu \}@zhejianglab.com.
 \IEEEcompsocthanksitem   Xiaogang Xu is with the Department of Computer Science and Engineering, The Chinese University of Hong Kong.
E-mail: xiaogangxu00@gmail.com. 
Ming-Hsuan Yang is with University of California at Merced.

\textbf{\faEnvelope{}}
Corresponding Author: Xiaogang Xu and Jiafei Wu. \protect\\}}

\markboth{Journal of \LaTeX\ Class Files,~Vol.~14, No.~8, April~2026}%
{Shell \MakeLowercase{\textit{et al.}}: A Sample Article Using IEEEtran.cls for IEEE Journals}

\maketitle

\begin{figure*}[t]
	\centering
  \vspace{-0.1in}
			\includegraphics[width=1.0\linewidth]{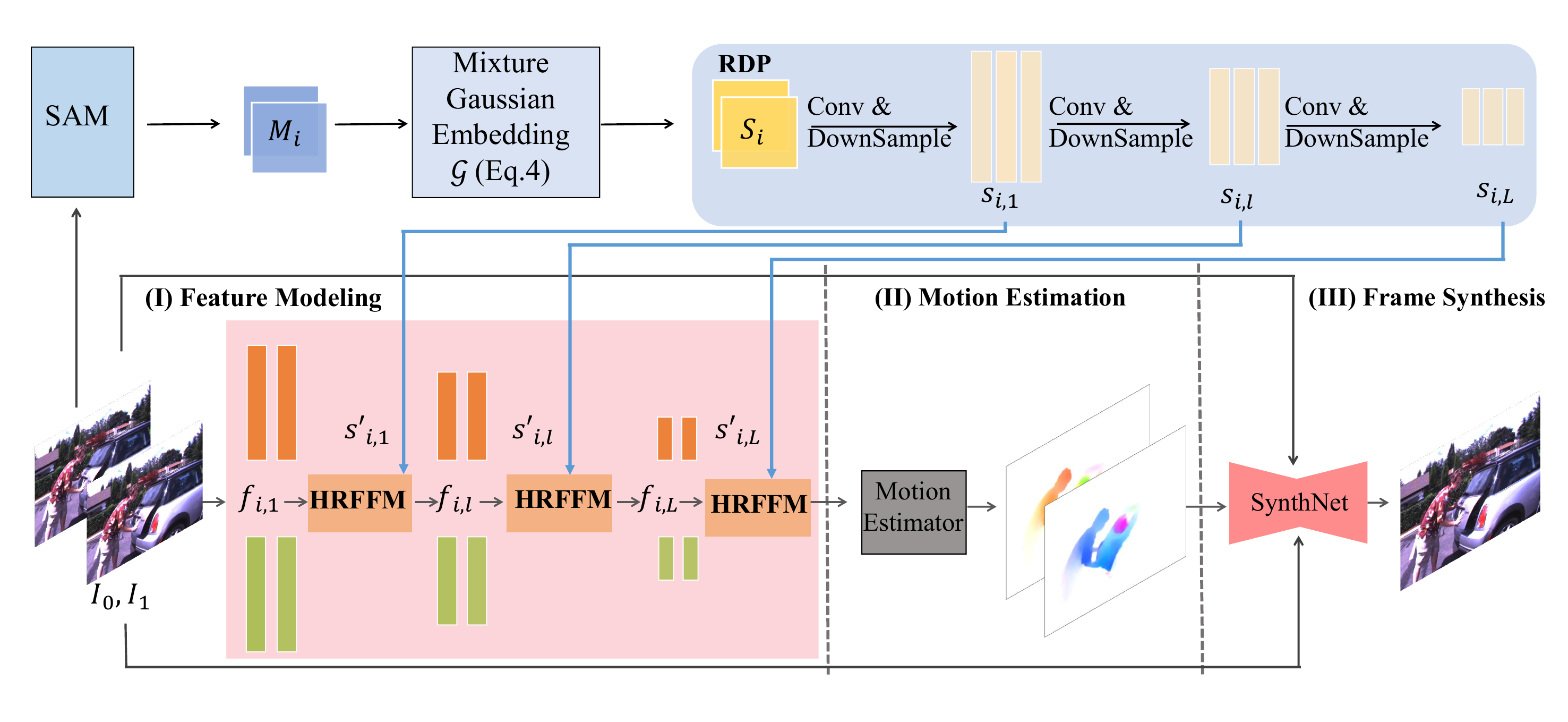}
		\vspace{-0.35in}
		\caption{The standard framework of motion-based VFI. It consists of three stages: extracting the image features from the encoder, making the optical flow estimation, and then warping and decoding it into a frame synthesis module to generate the intermediate frame. Our proposed HRFFM incorporates the prior RDP $S_i$ into the hierarchical stage of the encoder.}
		\label{fig:framework1}
		\vspace{-0.1in}
\end{figure*}

\begin{abstract}
In existing restoration-oriented Video Frame Interpolation (VFI) approaches, the motion estimation between neighboring frames plays a crucial role. However, the estimation accuracy in existing methods remains a challenge, primarily due to the inherent ambiguity in identifying corresponding areas in adjacent frames for interpolation. Therefore, enhancing accuracy by distinguishing different regions before motion estimation is of utmost importance. In this paper, we introduce a novel solution involving the utilization of open-world segmentation models, e.g., \textcolor{black}{SAM2 (Segment Anything Model2) for frames}, to derive Region-Distinguishable Priors (RDPs) in different frames. These RDPs are represented as spatial-varying Gaussian mixtures, distinguishing an arbitrary number of areas with a unified modality. RDPs can be integrated into existing motion-based VFI methods to enhance features for motion estimation, facilitated by our designed play-and-plug Hierarchical Region-aware Feature Fusion Module (HRFFM). HRFFM incorporates RDP into various hierarchical stages of VFI's encoder, using RDP-guided Feature Normalization (RDPFN) in a residual learning manner. With HRFFM and RDP, the features within VFI's encoder exhibit similar representations for matched regions in neighboring frames, thus improving the synthesis of intermediate frames. Extensive experiments demonstrate that HRFFM consistently enhances VFI performance across various scenes. 
\end{abstract}

\begin{IEEEkeywords}
VFI, Restoration, Region-Distinguishable Priors, Hierarchical Region-aware Feature Fusion.
\end{IEEEkeywords}

\section{Introduction}
\label{sec:intro}

Restoration-Oriented Video Frame Interpolation (VFI) represents a classic low-level vision task with the objective of augmenting video frame rates by generating intermediary frames that do not exist between consecutive frames. This technique has a wide range of practical applications, such as novel view synthesis~\cite{Alpher07}, video compression~\cite{Alpher08}, and cartoon creation~\cite{Alpher09}. Nevertheless, frame interpolation continues to present unsolved challenges, including issues related to occlusions, substantial motion, and alterations in lighting conditions. Enhancing the performance of existing restoration-oriented VFI frameworks (\textit{not generative models}) poses a significant challenge within both the research and industrial communities.

Restoration-oriented VFI methods can be broadly categorized into two main approaches: motion-free~\cite{Alpher33,ChannelM2020,Sepcheng2020,Alpher11} and motion-based~\cite{jiang2018super,liu2017video,Alpher14,Alpher15,softsplat2020,Alpher17,abme2021,Alpher19,Biformer,fitug,asymmetricble,zhang2023extracting,Vfimamba2024,KIM2025128728}, depending on whether they incorporate motion cues like optical flow. Motion-free models typically utilize methods such as kernel prediction or spatial-temporal decoding, which are effective while have limitations, such as being restricted to interpolating frames at fixed time intervals, and their runtime scales linearly with the number of desired output frames. On the other end of the spectrum, motion-based approaches establish dense correspondences between frames and employ warping techniques to generate intermediate pixels. Due to the explicit modeling of temporal correlations, motion-based strategies are more flexible.
Moreover, with recent advancements in optical flow technology~\cite{Alpher20,Alpher21,Alpher22,Alpher23}, motion-based interpolation's accuracy has been evolved.

Motion estimation between adjacent frames is a pivotal aspect of motion-based VFI. Nevertheless, achieving precise estimation accuracy in existing methods remains a formidable challenge, primarily due to the inherent ambiguity in identifying corresponding areas in adjacent frames for interpolation. This challenge becomes more pronounced when there is a substantial temporal gap in the target video. Previous research has predominantly focused on enhancing estimation accuracy by laboriously evolving network structures. In this paper, we posit that, in addition to network evolution, it is of paramount importance to enhance accuracy by differentiating between various regions prior to the motion estimation process.

In this paper, we present an innovative approach by introducing Region-Distinguishable Priors (RDPs) into motion-based VFI frameworks (\textcolor{black}{restoration-oriented, not generative models that have fidelity concerns}). These priors are derived from the existing open-source \textcolor{black}{Segment-Anything Model2 (SAM2)}~\cite{ravi2024sam2} with minimal impediments. \textcolor{black}{
\textit{In particular, SAM2 can be applied to videos to segment distinct regions across frames while maintaining consistent representations of these regions over time. The segmentation in each frame enables the estimation of appropriate, region-specific flow values for areas exhibiting different motions, and the temporal consistency of these representations facilitates more accurate motion estimation.
These properties satisfies our requirement.}
} 
Furthermore, we propose a new Hierarchical Region-aware Feature Fusion Module (HRFFM), which is designed to enhance the VFI framework's encoder, as illustrated in Fig.~\ref{fig:framework1}, to refine the corresponding features used in motion estimation. The HRFFM is a plug-and-play module that seamlessly integrates with various motion-based VFI methods without introducing a significant increase in network parameters.


The formulation of RDP from \textcolor{black}{SAM2} is not trivial, as RDP is required to differentiate objects with an arbitrary number, while the output of \textcolor{black}{SAM2} lacks a countable property. To make optimal use of the segment outputs from \textcolor{black}{SAM2} and provide them with the ability to distinguish multiple objects of the same dimensions, we have devised a novel Gaussian embedding strategy for the \textcolor{black}{SAM2} outputs. We employ the \textcolor{black}{SAM2} to produce instance segmentations for two input frames and utilize spatial-varying Gaussian mixtures to transform them into higher-dimensional RDPs. 
\textcolor{black}{Moreover, SAM2's temporally consistent segmentation ensures consistent RDP representations across frames.}
This representation has been demonstrated to outperform naive one-hot encoding or other learnable embedding alternatives.

The obtained temporally-consistent RDPs are integrated into the encoder of the target VFI model, with the primary goal of achieving regional consistency between neighboring frames in VFI. This means that the features of a specific region in two consecutive frames should be similar, which aids in the subsequent motion estimation process.
To achieve this objective, HRFFM incorporates RDP into the target model's hierarchical feature spaces and performs RDP-guided Feature Normalization (RDPFN) in a residual learning fashion to bring target features to desired states. RDPFN is novelly designed to simultaneously harness long- and short-range dependencies to fuse the RDP and image content, enabling the accurate estimation of regional normalization parameters.

\begin{figure*}[t]
	\centering
	\includegraphics[width=1.0\linewidth]{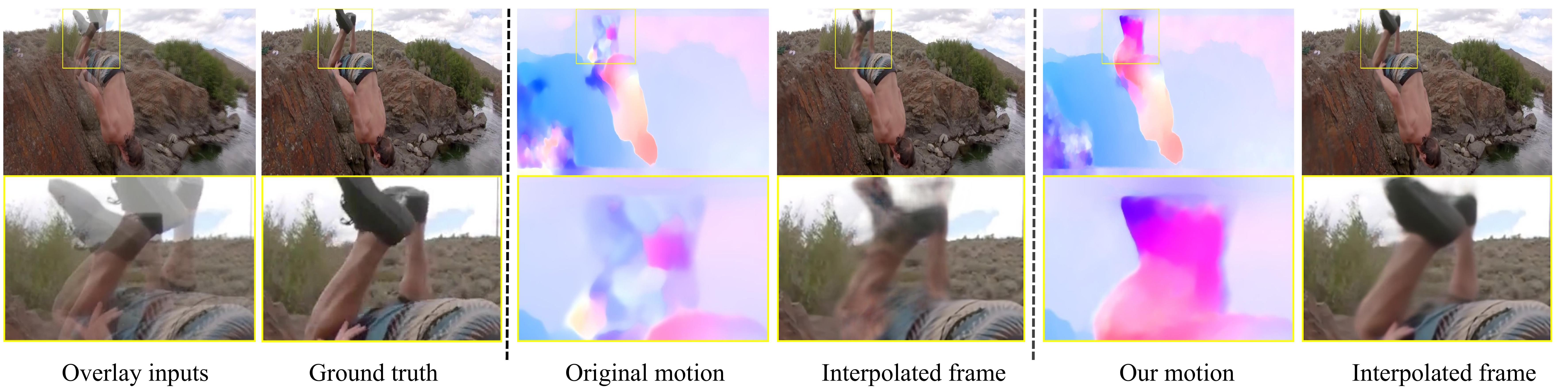}
		\vspace{-0.2in}
		\caption{The first two columns: overlay inputs and the ground truth frame. Middle two columns: motion field (from first to second frame) by VFIformer~\cite{Alpher19} and corresponding interpolation. The last two columns: motion field and interpolated frame by enhancing VFIformer with our strategy using RDPs. Our approach results in more satisfactory motion estimation and, thus, better interpolation results.}
		\label{fig:teaser}
        \vspace{-0.1in}
\end{figure*}

Extensive experiments are conducted on public and well-recognized datasets and various VFI networks. It's verified that our algorithm can bring stable performance improvement consistently on multiple datasets and models. Our strategy produces better motion modeling even with large motion scales, and thus enhances interpolated results (see Fig.~\ref{fig:teaser}).
\textcolor{black}{In particular, RDPs help produce clearer boundaries and mitigate severe degradations, such as blurring and shape incompleteness. This improvement stems from the fact that our RDPs are derived from SAM2, which provides accurate and well-defined mask boundaries, enabling the complete segmentation of a sufficient number of distinct instances.} \textcolor{black}{Meanwhile, the consistency of RDPs between consecutive frames is maintained.
}
In summary, our contribution is three-fold.
\begin{itemize}
    \item \textcolor{black}{We underscore the significance of distinguishing different regions within frames to enhance motion estimation and ultimately improve the performance of restoration-oriented VFI. To achieve this, we have innovatively devised a novel and systematic formulation pipeline for RDP using a Gaussian embedding strategy based on the output of SAM2.}
    \item \textcolor{black}{A new Hierarchical Region-aware Feature Fusion Module is designed to incorporate RDPs into the target model's encoder. It is a general and plug-and-play strategy for different networks.}
    \item Experimental results on different datasets and networks demonstrate the effectiveness of our proposed strategy. 
\end{itemize}

\textcolor{black}{Moreover, two key points should be clarified to highlight the role and applicability of our strategy.}

\noindent{\textbf{The relation between our strategy with existing works.}}
\textcolor{black}{Enhancing interpolation performance can follow two complementary roadmaps: enabling the network to distinguish various regions and evolving the network architecture itself. 
Updating neural networks alone can be effective, as studied by most existing works~\cite{Alpher19,Alpher39,Alpher40,li2023amt,seo2025bim}. However, this approach typically requires large model capacities and extensive training data to handle severe degradations.
This is why we propose a new roadmap by incorporating RDPs, which serve as effective priors to further boost the performance of existing VFI networks that have already undergone significant evolution—without introducing substantial additional parameters or requiring large-scale datasets.}

\noindent\textbf{The relation between our strategy with generative models.}
\textcolor{black}{Recently, some approaches have adopted generative models for frame interpolation~\cite{wangframer,zhu2025generative,wang2024generative,danier2024ldmvfi}. They primarily rely on pre-trained weights to synthesize frames rather than restore them, focusing on visual quality and aesthetics rather than fidelity. \textit{In contrast, our method is designed for restoration-oriented VFI, which emphasizes temporal alignment and leverages RDPs to enhance fidelity}. Given the fundamentally different objectives, our approach and generative-model-based strategies belong to distinct categories. Therefore, we have not applied our method to generative models.
Additionally, we have not trained or evaluated our method on datasets specifically designed for generative models, such as OpenVid~\cite{nan2024openvid}. Moreover, specific tasks like cross-dataset evaluation (e.g., between natural and cartoon videos) are also beyond the scope of our method, as they require strong generative capabilities.
}

\section{Related Work}
\subsection{Video Frame Interpolation}
\textcolor{black}{Except for diffusion-based models~\cite{lyu2025tlbvfitemporalawarelatentbrownian,danier2024ldmvfi,lyu2024fiwcbbd}, the current restoration-oriented VFI methods} can be broadly categorized into two groups: motion-free and motion-based approaches. Motion-free methods typically create intermediate frames by directly concatenating input frames. Such methods can be further classified into two types:  directly-generated methods~\cite{ChannelM2020,Alpher25,Alpher26,Alpher27} and kernel-based methods~\cite{Sepcheng2020,Alpher29,Alpher30,Alpher31,Alpher32,Alpher33,Alpher34,Alpher35}  concerning the generation of intermediate frames. Despite their simplicity, these methods lack robust modeling of motion, making it challenging to align corresponding regions between intermediate frames and input frames. This limitation often results in image blur and the presence of artifacts~\cite{Alpher36}.

Motion-aware methods explicitly model motion, often represented by optical flow, between two frames to enhance the alignment of distinguishable region information from input frames to intermediate frames. Some early approaches focused solely on predicting inter-frame motion for pixel-level alignment~\cite{jiang2018super,liu2017video,liu2019deep}.
\textcolor{black}{Subsequent works~\cite{Alpher19,Alpher14,softsplat2020,Alpher17,abme2021,Alpher41,sim2021xvfi,reda2022film,zhong2024clearer,liu2023ttvfi,liu2024sparse,liu2022jnmr,hu2023video,hu2024iq,Yoo_2023_ICCV,Biformer,fitug,asymmetricble,zhang2023extracting,Vfimamba2024,KIM2025128728}\textcolor{black}{~\cite{11414159,11339946,11268514}} have introduced separate modules for explicit motion modeling and motion refinement through synthesis, thereby enhancing overall performance.}
While the current state-of-the-art method has achieved impressive results, these systems still cannot handle practical challenges and need further performance improvement~\cite{zhang2023extracting}. Our proposed method offers a novel perspective by incorporating   Region Distinguishable Priors into motion-based VFI. Our designed play-and-plug Hierarchical Region-aware Feature Fusion Module provides a straightforward and efficient approach to improving VFI features via RDPs.

\textcolor{black}{
Moving beyond traditional optical flow, recent studies ~\cite{10659365}~\cite{Xiao_2024} prioritize efficient temporal difference modeling to reduce computational overhead. Methods like the Temporal Difference Module (TDM) for satellite VSR and Sparse Spatio-Temporal Propagation (SSTP) for unified segmentation utilize sparse 3D convolutions and temporal discrepancies to achieve robust compensation without explicit flow estimation. Similarly, advanced mechanisms in Multi-Axis Feature Diversity Enhancement~\cite{10918606} and CycMuNet+~\cite{10177211} highlight the value of sophisticated motion modeling. Although distinct from Video Super-Resolution (VSR) which recovers spatial details, our Video Frame Interpolation (VFI) task benefits from these shared principles to effectively synthesize non-existent temporal states, confirming the necessity of robust spatio-temporal feature representation.}

\subsection{\textcolor{black}{Segmentation Models}}

\textcolor{black}{Traditional segmentation models can be broadly categorized into three types: semantic segmentation, instance segmentation, and panoptic segmentation. The goal of semantic segmentation~\cite{zhao2017pyramid,chen2017deeplab} is to assign a predefined category label to every pixel in an image. However, it does not distinguish among different instances of the same class— all objects belonging to the same category are labeled identically.
To address this limitation, instance segmentation was introduced~\cite{wang2021solo,lee2020centermask}, which not only classifies each pixel but also differentiates among individual object instances of the same category.
Building upon this, panoptic segmentation~\cite{kirillov2019panoptic} was proposed as a unified framework that combines semantic and instance segmentation. It separates image content into ``Thing" classes (countable objects, such as people or cars) and ``Stuff" classes (amorphous regions, like sky or road).
Despite these advances, both instance and panoptic segmentation are limited in scope: not all regions in an image can be meaningfully or distinctly segmented with an instance-aware mask.}

Recently, the foundational Computer Vision (CV) model for Segment Anything, known as SAM~\cite{kirillov2023segment}, was recently unveiled. 
SAM is a substantial Vision Transformer (ViT)-based model that underwent training on an extensive visual corpus (SA-1B). 
\textcolor{black}{Despite lacking semantic ability, SAM's} capabilities in segmentation have shown promise across various scenarios, underscoring the significant potential of foundational models in the realm of CV. 
This development marks a groundbreaking stride toward achieving visual artificial general intelligence. 

SAM has shown its versatility across several tasks, extending its assistance beyond segmentation. Tasks such as image synthesis~\cite{Alpher37} and video super-resolution~\cite{Alpher38} have all benefited from SAM's capabilities. \textcolor{black}{Segment Anything Model 2 (SAM2) \cite{ravi2024sam2} is a foundation model designed for promptable visual segmentation in both images and videos, extending SAM by treating images as single-frame videos. More recently, Segment Anything Model 3 (SAM3) has been introduced as a unified framework that detects, segments, and tracks objects in images and videos based on concept prompts. The continuous evolution of SAM models has greatly advanced the use of segmentation-based priors in real-world videos.} In a pioneering effort, we've explored SAM2's potential in VFI, marking the first attempt to apply SAM2 to this domain. Extensive experiments substantiate that SAM2 significantly enhances VFI effectiveness.

\textcolor{black}{
In this paper, we use SAM2 to obtain RDP priors instead of other segmentation models, based on a comparative analysis. Semantic segmentation models cannot distinguish between different instances of the same category—for example, two cars moving in opposite directions receive the same label, failing to capture motion differences. Although instance and panoptic segmentation models can separate instances, they are limited by a fixed set of predefined categories and may miss objects not covered by those categories, thus failing to capture all motion patterns. In contrast, SAM2 can segment nearly all objects in an image and distinguish an arbitrary number of instances, making it more suitable for our task.
Experiments will further highlight the advantage of using SAM2.  
}

\section{Method}
\label{sec:sec3}

In this section, we first provide the overview of our strategy in Sec.~\ref{sec:overview}. Then, two vital components in our framework, i.e., the formulation of Region-Distinguishable Priors and the design of HRFFM, will be elaborated in Sec.~\ref{sec:sec31} and \ref{sec:sec32}, respectively. One significant component in HRFFM, i.e.,  RDP-guided Feature Normalization (RDPFN), will be introduced in Sec.~\ref{sec:sec33}. Sec.\ref{sec:defr} provides a detailed explanation of the architectures of our networks.

\subsection{Overview}
\label{sec:overview}

\noindent\textbf{Task setting.}
Given two frames $I_0,I_1 \in \mathbb{R}^{H\times W \times 3}$, the target of VFI is to synthesize an intermediate frame ${\hat{I}_{t}\in {\mathbb{R}}^{H\times W\times 3}}$ at arbitrary time step $t\in(0,1)$, as
\begin{equation}{
\hat{I}_{t}=\mathcal{O}(I_0,I_1,t),
}
\label{eq:1}
\end{equation}
where $\mathcal{O}$ denotes the VFI method that shares a common framework as illustrated in Fig.~\ref{fig:framework1}. Motion-based VFI typically comprises three key stages. These stages involve feature extraction for $I_0$ and $I_1$, with the extracted features labeled as $f_{0,l}$ and $f_{1,l}$, where $l \in [1, L]$ signifies the $l$-th layer in the encoder. Additionally, it includes motion estimation between the extracted features and warping these features to synthesize the final results. The accuracy of the motion estimation stage holds pivotal importance within VFI, as it directly influences the ultimate performance.

\noindent\textbf{Challenge.}
While numerous motion estimation strategies have been introduced in recent years, their effectiveness is predominantly evident in scenarios involving continuous motions. However, in the context of VFI tasks, there exists a substantial temporal gap and limited continuity between adjacent frames. This presents a significant challenge for accurate motion estimation. The primary obstacle in this motion estimation process arises from the inherent ambiguity associated with identifying corresponding areas in neighboring frames for interpolation. Consequently, achieving precise estimation accuracy in current VFI frameworks remains a formidable challenge

\noindent\textbf{Motivation.}
To address the aforementioned challenge, we propose a method to enhance the extracted features for interpolation by introducing specific priors capable of distinguishing different objects within frames. This serves to reduce ambiguity in the identification of matching areas in adjacent frames. These priors are obtained through the utilization of the current open-world segmentation module, such as SAM, resulting in $M_0$ and $M_1$ for $I_0$ and $I_1$. Furthermore, these priors are integrated hierarchically into the feature extraction stage of VFI models, given that VFI models typically employ pyramidal structures in their encoders. The primary objective is to provide distinct feature representations for different areas within $I_0$ and $I_1$. This, in turn, enables more accurate motion estimation by distinguishing between various objects and being aware of boundaries.

\noindent\textbf{Implementation.}
Given $I_0$ and $I_1$, we first obtain their SAM2 outputs as $M_0$ and $M_1$. Then, $M_0$ and $M_1$ (${M_0,M_1}\in \mathbb{R}^{H\times W\times 1}$) are transformed into the desired Region-Distinguishable Priors (RDPs) that can distinguish different regions in frames with a unified representation dimension. 
Thus, Eq.~\ref{eq:1} can be written as 
\begin{equation}{
\hat{I}_{t}=\mathcal{O}(I_0,I_1, \mathcal{G}(M_0),\mathcal{G}(M_1),t),
}
\label{eq:eq11}
\end{equation}
where $\mathcal{G}$ is the transformation function to produce RDPs, and we denote $\mathcal{S}_0=\mathcal{G}(M_0)$ and $\mathcal{S}_1=\mathcal{G}(M_1)$.
The extracted features $f_{0,l}$ and $f_{1,l}$ are enhanced with our proposed Hierarchical Region-aware Feature Fusion Module (HRFFM) (as displayed in Fig.~\ref{fig:framework1}), as
\begin{equation}
f'_{0,l}=\mathcal{H}(f_{0,l}, \mathcal{S}_0), \; f'_{1,l}=\mathcal{H}(f_{1,l}, \mathcal{S}_1),
\end{equation}
where $\mathcal{H}$ is the designed HRFFM.
The enhanced $f'_{0,l}$ and $f'_{1,l}$ are then sent to the following original motion estimation and frame synthesis stages to obtain the final result.

\subsection{Region-Distinguishable \textcolor{black}{Priors (RDPs)}}
\label{sec:sec31}

\noindent\textbf{The drawback of SAM2 outputs for VFI.}
The original SAM2 model provides segmentation outputs for all instances within an image. SAM2 generates masks for frames, with each pixel value representing an object. Its remarkable segmentation capabilities make it a valuable choice as a region-distinguishable prior. Over time, several variants of SAM2 have been introduced, enhancing its capabilities, including semantic and panoptic segmentation when combined with other models. However, SAM2's output has limitations when it comes to representing objects with arbitrary numbers, a requirement for RDP. The semantic one-hot embedding is constrained by semantic categories, and the instance one-hot embedding assumes a maximum instance number, making it unable to accommodate new instances during real-world evaluation. Consequently, there is a need to transform SAM2's output to make it more suitable for RDPs.

\noindent\textbf{Mixture Gaussian embedding strategy.}
We posit that the representations of segmentation priors can be conceptualized as distributed sampling results with distinct parameters across different regions of an image. These parameters enable the discrimination of regions and the alignment of the same region across multiple frames. In particular, each segmented area can be interpreted as a sampling result from a Gaussian distribution characterized by individual parameters. To facilitate this corresponding sampling process, we begin by establishing a codebook $\mathcal{C}$ that comprises a range of Gaussian parameters, encompassing both mean and variance. Subsequently, each object identified by the SAM2 output can retrieve its specific Gaussian parameters via a hashing mechanism.

\textcolor{black}{We elaborate the procedure and characteristic of the hashing mechanism.
First, each region will be assigned an index by SAM2, which can be used to look up the corresponding location in $\mathcal{C}$ with the same index.
Then, we can assign unique mean and variance values from $\mathcal{C}$ to each region.
In particular, we aim to set a maximum value that accommodates all possible object counts within the frames. This ensures that each region corresponding to an object has a unique entry in the codebook, avoiding collisions with other regions. 
In this way, not all codebook elements are utilized to generate RDPs; instead, the number of elements used varies across different images, not fixed, depending on the objects present. }

In summary, the transformation procedure can be written as:
\begin{equation}{
{S}_{i}=\mathcal{G}(M_i)=\mathcal{N}(\mathcal{C}_m(M_i), \mathcal{C}_v(M_i)), \; i=0,1,
}
\end{equation}
where ${S_i}\in \mathbb{R}^{H\times W\times c}$, $\mathcal{N}$ is the Gaussian distribution sampler, $\mathcal{C}_m$ is the codebook for Gaussian mean values, and $\mathcal{C}_v$ is the codebook for Gaussian variance scores.
This Gaussian mixture is independent of the number of object types (adding a new object in the frame equals sampling a new Gaussian parameter), and distinguishes an arbitrary number of areas with a unified modality.
\textcolor{black}{To provide a rigorous foundation for our mask transformation, we have formalized the process as a High-dimensional Continuous Feature Mapping via Gaussian Random Projection and Spatial Interpolation.}

\textcolor{black}{Let the SAM2 output mask be $M \in \mathbb{N}^{H \times W}$, where each element $M(i, j) \in \{1, 2, \dots, K\}$ represents a discrete class ID (or object ID) at pixel $(i, j)$. Our goal is to transform this into a continuous prior ${S} \in \mathbb{R}^{H \times W \times c}$, where $c$ is the target dimensionality. We first initialize a Gaussian Random Projection Matrix $\mathcal{G}\in \mathbb{R}^{K\times c}$, where each row $g_{k}\in \mathbb{R}^{c}$ is independently sampled from a multivariate Gaussian distribution. By leveraging the properties of high-dimensional random mapping, the distinctness between different regions is preserved in the embedding space. This process effectively converts the non-differentiable categorical IDs into a structured, continuous representation that is more compatible with the subsequent deep neural layers for spatial reasoning and feature fusion.}

\textcolor{black}{Note that the codebook entries are pre-defined and not learned. Their primary role is to provide distinct values to represent different objects, which are used to generate the RDPs. The learning process is carried out by the subsequent HRFFM network, which learns to effectively utilize these RDPs through feature fusion. Further details are provided in the next section.}

\begin{figure*}[t]
  \centering
   \includegraphics[width=0.6\linewidth]{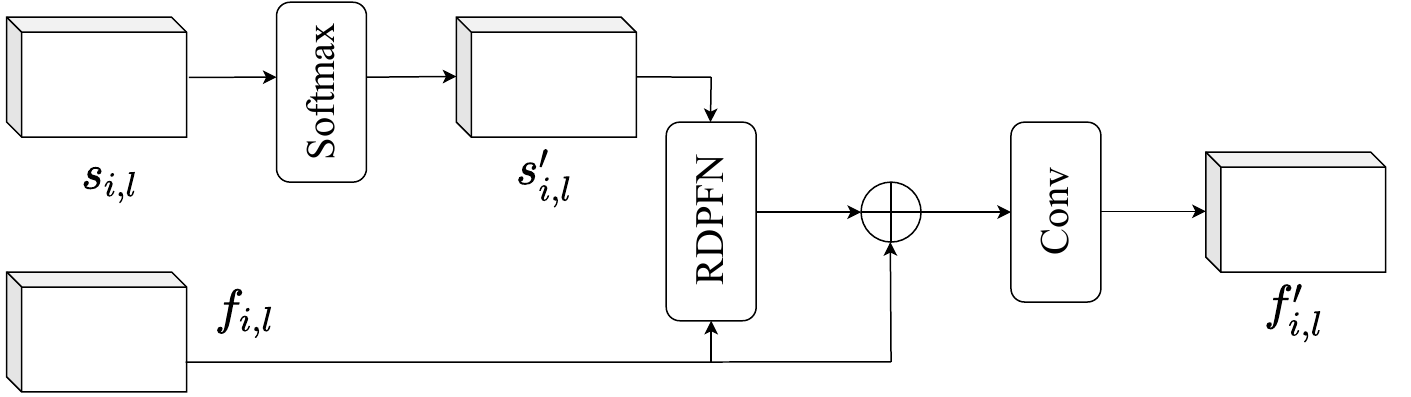}
   \caption{The Overview of HRFFM, which first exploits RDPs to enhance image features via RDPFN (Eq.~\ref{eq:fuse}), and then refine it via refinement (Eq.~\ref{eq:residual}). ${f_{i,l},s_{i,l}}$ are the image feature and RDP feature of the $i$-th frame of the $l$-th layer, respectively. $\oplus$ means concatenating. }
   \label{fig:fig3}
\end{figure*}

\subsection{HRFFM}
\label{sec:sec32}

As indicated in Sec.~\ref{sec:overview}, standard motion-based VFI conducts multi-scale feature extraction before motion estimation. Thus, we put the obtained RDPs (from Sec.~\ref{sec:sec31}) into each layer of image feature extraction as shown in Fig.~\ref{fig:framework1}. The fusion consists of three stages, including RDP feature extraction, RDP-guided Feature Normalization (RDPFN), and RDP residual learning, as exhibited in Fig.~\ref{fig:fig3}.

To seamlessly integrate RDP into different layers of the target VFI, we must perform feature extraction for RDP in a pyramidal fashion, resulting in the acquisition of $s_{i,l}$, where $i\in{0,1 }$ and $l\in[1,L]$, from $\mathcal{S}_i$. This approach ensures that $s_{i,l}$ and $f_{i,l}$ share the same shape in the deep feature space, facilitating their fusion. Furthermore, it's imperative to unify $s_{i,l}$ at each layer into a region-distinguishable distribution to prevent inconsistencies among different layers.
To this end, the RDP input of each layer is written as
\begin{equation}
    {s'_{i,l}}=\mathcal{M}(s_{i,l}),
\end{equation}
where $\mathcal{M}$ is the softmax operation.

In order to enhance the distinctiveness of features across different regions and improve the precision of matching during the motion estimation stage, we have introduced RDP-guided Feature Normalization (RDPFN). RDPFN takes inputs in the form of $f_{i,l}$ and ${s_{i,l}}^{'}$, and it produces region-aware feature normalization parameters. The resulting normalized feature is denoted as $\hat{f}_{i,l}$, as
\begin{equation}
    \hat{f}_{i,l}=\mathcal{R}_l(f_{i,l}|{s'_{i,l}}),
    \label{eq:fuse}
\end{equation}
where $\mathcal{R}_l$ is the RDPFN operation in the $l$-th layer. The details of RDPFN will be introduced in Sec.~\ref{sec:sec33}.

\begin{figure}[t]
		\centering
			
            \includegraphics[width=1\linewidth]{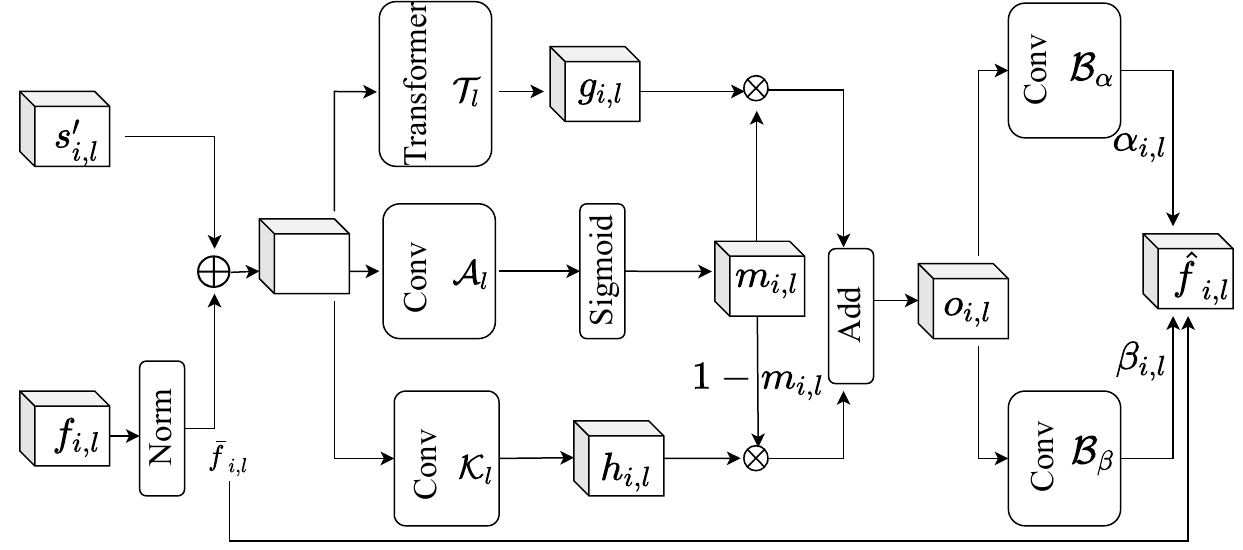}
		\caption{The Overview of RDPFN. It utilizes both RDP features and image features as inputs. It employs a combination of long- and short-range operations to extract impactful features, facilitating the prediction of region-aware normalization parameters. This approach ensures that features within the same instance exhibit similarity, thereby enhancing the effect of subsequent modules. ${\oplus}$ means concatenating , ${\otimes}$ means dot producting.}
		\label{fig:fig4}
\end{figure}

Moreover, we recognize that segmentation results obtained from \textcolor{black}{SAM2} may contain errors when dealing with diverse real-world images. Consequently, additional refinement operations are essential to enhance the features derived from RDPFN, rendering them more adaptable for subsequent motion estimation and frame synthesis. In our study, we have identified a refinement operation that enhances robustness and is accomplished through a spatial-channel convolution fusion in a residual manner, as
\begin{equation}
    {f'_{i,l}}=\mathcal{V} (\hat{f}_{i,l}, f_{i,l}),
    \label{eq:residual}
\end{equation}
where $\mathcal{V}$ denotes the convolution operation for fusion.

\subsection{RDP-guided Feature Normalization}
\label{sec:sec33}

To fuse $f_{i,l}$ and ${s'_{i,l}}$ in Eq.~\ref{eq:fuse}, RDPFN will predict the region-aware feature normalization parameters, making different areas to be distinguishable in the deep feature space. The normalization parameters contain the scaling parameter $\alpha_l$ and the bias parameter $\beta_l$.

The input to RDPFN includes both image features, represented as $f_{i,l}$, and RDP, denoted as ${s'_{i,l}}$. This is because image features play a crucial role in identifying corresponding areas in neighboring frames with similar appearances. The synergy of image features and RDP enables the discovery of instance-level matched regions.

To derive the appropriate normalization parameters, we employ a flexible and lightweight backbone capable of capturing information from both local and global perspectives. This choice is intuitive since certain regions, characterized by small areas, benefit from local information for more accurate discrimination, while larger regions necessitate long-range information. As illustrated in Fig.~\ref{fig:fig4}, our backbone consists of parallel CNN and transformer blocks, denoted as $\mathcal{T}_l$ and $\mathcal{K}_l$, respectively. Differing from the conventional CNN-transformer structure, we introduce a learnable fusion mask, denoted as $m_{i,l}$ that is predicted by $\mathcal{A}$.

The overall pipeline can be denoted as the following equations, as
\begin{equation}
    \begin{aligned}
    &\bar{f}_{i,l}=\mathrm{Norm}(f_{i,l}),\\
        &g_{i,l}=\mathcal{T}_l(\bar{f}_{i,l} \oplus {s'_{i,l}}), \; h_{i,l}=\mathcal{K}_l(\bar{f}_{i,l} \oplus {s'_{i,l}}),\\
        &m_{i,l}=\mathrm{Sigmoid}(\mathcal{A}_l(\bar{f}_{i,l} \oplus {s'_{i,l}})),\\
        &o_{i,l}=g_{i,l} \times m_{i,l} + h_{i,l} \times (1-m_{i,l}),\\
        & \alpha_{i,l}=\mathcal{B}_\alpha(o_{i,l}), \; \beta_{i,l}=\mathcal{B}_\beta(o_{i,l}),\\
        & \hat{f}_{i,l}=\bar{f}_{i,l} \times (1+\alpha_{i,l}) + \beta_{i,l},
    \end{aligned}
    \label{eq:pipeline}
\end{equation}
where $\mathrm{Norm}$ is the ordinary normalization operation, $\mathrm{Sigmoid}$ is the sigmoid activation function, $\mathcal{B}_\alpha$ and $\mathcal{B}_\beta$ are two light-weight convolution layers to obtain the normalization parameter prediction results, $\hat{f}_{i,l}$ is the output feature from RDPFN as shown in Eq.~\ref{eq:fuse}.

\subsection{Details of Our Framework's Components}
\label{sec:defr}
Our framework comprises two key components: 1) Region-Distinguishable Priors (RDPs), represented as spatially-varying Gaussian mixture models, and 2) the core Hierarchical Region-aware Feature Fusion Module (HRFFM), enhancing feature representations for accurate motion estimation.

RDPs come out of the generated mask $M_i$ from SAM, and a Gaussian distributed sampler $\mathcal{G}$, written as ${S}_{i}=\mathcal{G}(M_i)$, where ${S_i}\in \mathbb{R}^{H\times W\times c}$.
In our experiments, $c$ is 3, the same dimension as the input frame. An ablation study to analyze the impact of this channel number will be given in Sec.~\ref{sec:absd}, verifying that 3 is the optimal hyper-parameter.

The HRFFM consists of three parts, including RDP feature extraction, RDP-guided Feature Normalization (RDPFN), and RDP residual learning. In the procedure of RDP feature extraction, two multi-scale encoders are set for RDPs and input frames, respectively. They both have a series of convolution layers, guaranteeing the same size of frame/RDP encoder's outputs. HRFFM can be adopted as a play-and-plug module for different VFI frameworks, and we take UPR-Net as an example here.
As shown in Tab.~\ref{table:rdp-feature}, the feature encoder of UPR-Net has three convolutional stages: stage-0, stage-1, and stage-2. Each stage consists of four convolution layers, and the first layers in stage-1 and stage-2 perform down-sampling. Analogously, the RDP feature encoder, UPR-Net$_{ours}$, has three convolution stages. Each stage consists of two convolution layers as shown in Tab.~\ref{table:rdp-feature2}, and the first layers of stage-1 and stage-2 perform down-sampling. One difference is that the RDP feature extraction discards activation functions to speed up training.

In RDPFN, the $l$-th layer feature of frame $f_{i,l}$ is first transformed to $\bar{f}_{i,l}$ by a batch normalization operation.
Then we concatenate it with the $l$-th layer feature of RDP $s'_{i,l}$ to get $c_{i,l}$-D embeddings (where $c_{i,l}=c_{s'_{i,l}} + c_{\bar{f}_{i,l}}$, $c_{s'_{i,l}}$ and $c_{\bar{f}_{i,l}}$ represent the feature channels of $s'_{i,l}$ and $\bar{f}_{i,l}$, respectively), and send it to: (1) the CNN block $\mathcal{K}_l$ that consists of two convolution layers (the strides are both set as 1) with output channel number as $c_{i,l}$; (2) the Transformer block $\mathcal{T}_l$ whose procedure is written as:
\begin{equation}
    \begin{aligned}
       &x'_{i,l}=x_{i,l}+\rm{Atten}(Norm(x_{i,l})),\\ 
       &x''_{i,l}=x'_{i,l}+\rm{FeedForward}(Norm(x'_{i,l})),
    \end{aligned}
    \label{eq:attention_norm}
\end{equation}
where $x_{i,l}$ is the input feature of $\mathcal{T}_l$,  ``Atten'' is the multi-head self-attention (the number of heads is 2), and ``FeedForward'' is the feed-forward module in the transformer.
Moreover, the learnable fusion $\mathcal{A}_l$, normalization  prediction networks $\mathcal{B}_\alpha$ and $\mathcal{B}_\beta$ are both convolution layers with the stride of 1. The output of RDPFN $\hat{f}_{i,l}$ concatenates with $f_{i,l}$ and goes through a spatial
channel convolution fusion to achieve the final augmented frame features $f'_{i,l}$.

\begin{table}[t]
    \setlength{\belowcaptionskip}{0pt}
    \renewcommand\arraystretch{1.8}
    \centering
    \caption{Feature encoder of input frames in UPR-Net \cite{Alpher39}.}
    \resizebox{0.99\columnwidth}{!}{
    \begin{tabular}{|l|c|c|c|c|c|c|}
    \hline
         &Layer Type  & Activation & Kernel & Stride & Padding &Output Size\\
    \hline
    \hline
   \multirow{2}{*}{stage-0}&Input Fearure & -&- & -& -& $H\times W \times 3$\\
\cline{2-7}&Convolution*4&LeakyReLu & 3&1&1&$H\times W \times 16$\\
\hline
\multirow{2}{*}{stage-1}&Convolution&LeakyReLu & 3&2&1&$\frac{H}{2}\times \frac{W}{2} \times 32$\\
\cline{2-7}
&Convolution*3&LeakyReLu & 3&1&1&$\frac{H}{2}\times \frac{W}{2} \times 32$\\
    \hline
\multirow{2}{*}{stage-2}&Convolution&LeakyReLu & 3&2&1&$\frac{H}{4}\times \frac{W}{4} \times 64$\\
\cline{2-7}
&Convolution*3&LeakyReLu & 3&1&1&$\frac{H}{4}\times \frac{W}{4} \times 64$\\
    \hline
    \end{tabular}}
    
    \label{table:rdp-feature}
    
\end{table}

\begin{table}[t]
    \setlength{\belowcaptionskip}{0pt}
    \renewcommand\arraystretch{1.8}
    \centering
    \caption{The feature encoder of RDPs in UPR-Net$_{ours}$.}
    
    \resizebox{0.99\columnwidth}{!}{
    \begin{tabular}{|l|c|c|c|c|c|c|}
    \hline
        & Layer Type  & Activation & Kernel & Stride & Padding &Output Size\\
    \hline
    \hline
  \multirow{2}{*}{stage-0} &RDPs Fearure & -&- & -& -& $H\times W \times 3$\\
   \cline{2-7}&
Convolution*2& - & 3&1&1&$H\times W \times 16$\\
\hline
\multirow{2}{*}{stage-1}&Convolution&- & 3&2&1&$\frac{H}{2}\times \frac{W}{2} \times 32$\\
\cline{2-7}&
Convolution&- & 3&1&1&$\frac{H}{2}\times \frac{W}{2} \times 32$\\
    \hline
\multirow{2}{*}{stage-2}&Convolution&- & 3&2&1&$\frac{H}{4}\times \frac{W}{4} \times 64$\\
\cline{2-7}&
Convolution&- & 3&1&1&$\frac{H}{4}\times \frac{W}{4} \times 64$\\
    \hline
    \end{tabular}}
    
    \label{table:rdp-feature2}
    
\end{table}

\textbf{We will release all the code and models upon the publication of this paper}.

\section{Experiments}
\label{sec:sec4}

\subsection{Datasets}

Our model is trained on the Vimeo90K training set and
evaluated on various datasets.

\noindent\textbf{Training dataset.} 
The Vimeo90K dataset \cite{Alpher41} contains 51,312 triplets with a resolution of 448$\times$256 for training.
We augment the training images by randomly cropping
256$\times$256 patches. We also apply random flipping, rotating, and reversing the order of the triplets for augmentation. \textcolor{black}{ 
 Except for the results on the SNU-FILM Extreme subset in Tab.~\ref{tab:tab1} which were obtained by further training on the Easy, Medium, and Hard subsets, all other results were trained exclusively on the Vimeo90K dataset.
 }

\noindent\textbf{Evaluation datasets.} 
While these models are exclusively trained on Vimeo90K, we assess their performance across a diverse range of benchmarks featuring various scenes.
\begin{itemize}
\item \textbf{UCF101~\cite{Alpher42}}: The test set of UCF101 contains 379 triplets with a resolution of 256$\times$256. UCF101 contains a large variety of human actions.
\item \textbf{Vimeo90K~\cite{Alpher41}}: The test set of Vimeo90K contains 3,782 triplets with a resolution of 448$\times$256.
\item \textbf{SNU-FILM~\cite{ChannelM2020}}: This dataset contains 1,240 triplets, and most of them are of a resolution of around 1280$\times$720. It contains four subsets with increasing motion scales – easy, medium, hard, and extreme.

\end{itemize}

\noindent
\textbf{Metrics.} 
PSNR, SSIM~\cite{psnr}, \textcolor{black}{FID~\cite{2017GANs}}, LPIPS~\cite{zhang2018perceptual}, and NIQE~\cite{mittal2012making} are used for quantitative evaluation of frame interpolation. For running time, we test all models under 640$\times$480 resolution, and average the running time by 100 iterations. \textcolor{black}{Moreover, FVD is also employed~\cite{unterthiner2019fvd}}.

\begin{figure*}[t]
    \begin{center}
	\includegraphics[width=0.8\linewidth]{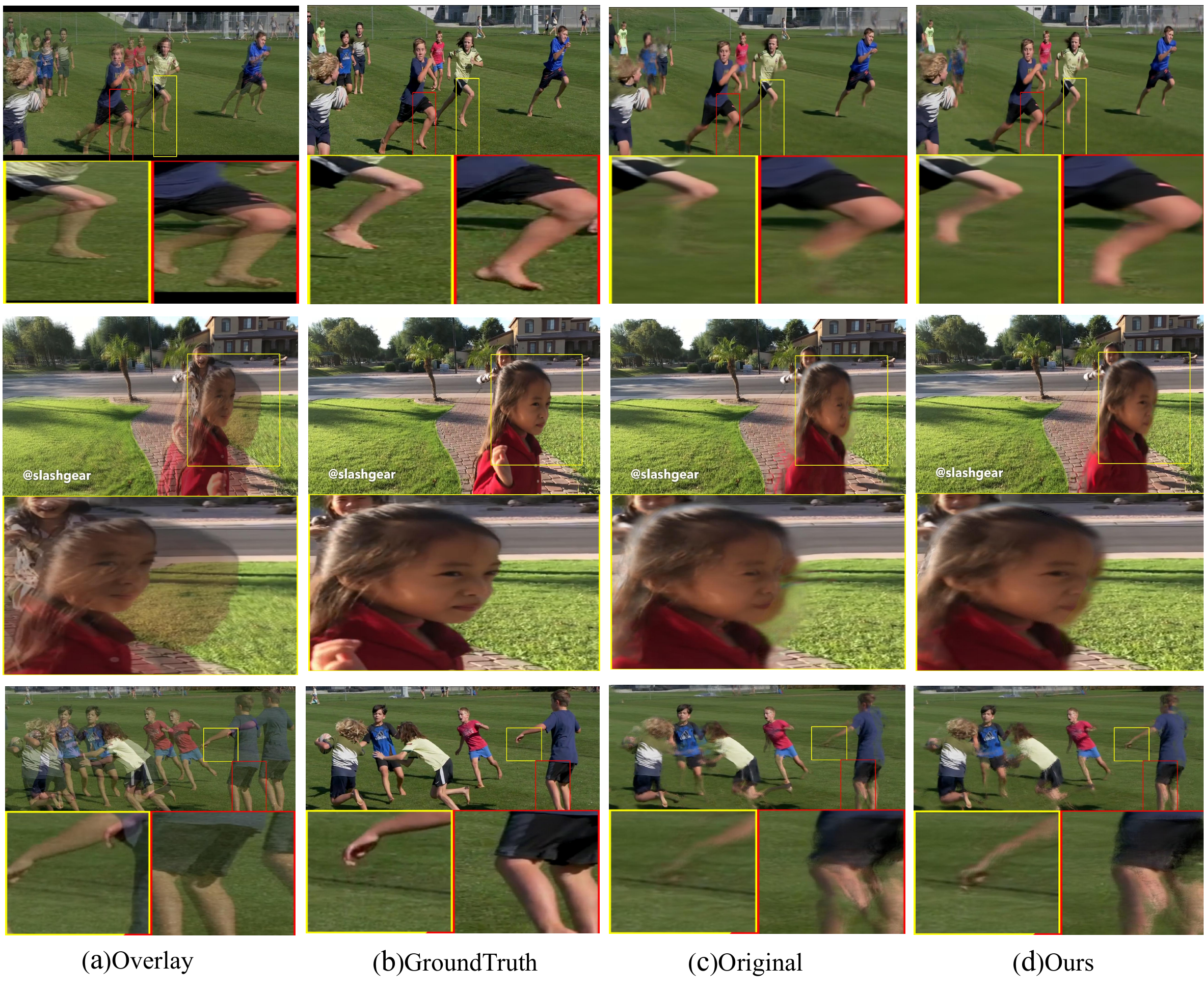}
	\end{center}
	\vspace{-0.25in}
	\caption{Visual comparison on SNU-FILM \cite{ChannelM2020}.
Three rows, from top to bottom, represent the comparison results for VFIformer, UPR-Net, and M2M-PWC. The highlighted boxes indicate positions where our model demonstrates superior performance.
}	\label{fig:rescompare}
     \end{figure*}

\begin{table}[t]
    \centering
    \caption{\textcolor{black}{Quantitative (PSNR/SSIM\textcolor{black}{/FID}/LPIPS) comparisons between VFI baselines and their implementation with our strategy ($_{ours}$) on  Vimeo90K~\cite{Alpher41}, and
the ``extreme" subset of SNU-FILM~\cite{ChannelM2020} benchmarks. 
Our strategy enhances the performance of various representative VFI methods. Note that our approach does not introduce a significant increase in computation cost. For running time, we test all models under 640 × 480 resolution, and average the running time by 100 iterations. Moreover, the results on Vimeo90K were obtained by training solely on the Vimeo90K dataset, while the SNU-FILM Extreme results were achieved by further fine-tuning the model on the training set augmented with the Easy, Medium, and Hard subsets, highlighting its improvement in challenging scenarios.
}}
\large
    \label{tab1}
    \resizebox{1\linewidth}{!}{\begin{tabular}{|l|c|c|c|c|}
    \hline
\multirow{2}{*}{Methods} & \multicolumn{1}{c|}{Vimeo90K} &
\multicolumn{1}{c|}{SNU-FILM Extreme} & parameters & runtime \\
\cline{2-5}
& PSNR$\uparrow$\,/\,SSIM$\uparrow$ &  PSNR$\uparrow$\,/\,SSIM$\uparrow$\,/\,FID$\downarrow$\,/\,LPIPS$\downarrow$ & (millions) & (seconds)\\
\hline
\hline
UPR-Net~\cite{Alpher39}	&36.02\,/\,0.9800& 25.67\,/\,0.8697\,/\,49.23\,/\,0.1452&
1.65&0.05 \\
\textbf{UPR-Net$_{ours}$}	&\textbf{36.19\,/\,0.9806}&		\textbf{25.94\,/\,0.8749\,/\,47.77\,/\,0.1382}&2.64&	 	0.13 \\
\hline
M2M-PWC~\cite{Alpher40} &	35.27\,/\,0.9771&	25.26\,/\,0.8663\,/\,62.34\,/\,0.3920 &7.61&	 	0.04 \\
\textbf{M2M-PWC$_{ours}$}&	\textbf{35.37\,/\,0.9775}&		\textbf{25.47\,/\,0.8682\,/\,57.32\,/\,0.3873}&10.65&	 	0.04 \\

\hline
AMT~\cite{li2023amt} & 36.50\,/\,0.9816 &	25.72\,/\,0.8723\,/\,54.14\,/\,0.2633 & 29.22 &	0.06 

 \\
\textbf{AMT$_{ours}$} &	\textbf{36.61\,/\,0.9828} &	\textbf{25.87\,/\,0.8772\,/\,48.99\,/\,0.2576} & 36.74 &	0.14 
 \\
\hline
Bim-VFI~\cite{seo2025bim}	& 35.12\,/\,0.9753 & 24.63\,/\,0.8551\,/\,43.12\,/\,0.2336 & 6.56 &	0.05 

 \\
\textbf{Bim-VFI$_{ours}$}	&\textbf{35.31\,/\,0.9763
}&	\textbf{24.75\,/\,0.8573\,/\,40.37\,/\,0.2308} & 10.82 &	0.06 
\\

\hline

PerVFI~\cite{asymmetricble}	& 33.60\,/\,0.9256 &24.02\,/\,0.7560\,/\,49.36\,/\,0.1230 &
8.67 &0.09 \\
\textbf{PerVFI$_{ours}$}	&\textbf{33.94\,/\,0.9578}&		\textbf{24.88\,/\,0.7890\,/\,37.52\,/\,0.1000}& 11.28 &	0.19 \\
\hline
\end{tabular}}
\label{tab:tab1}
\end{table}

\begin{table*}[t]
    \centering
    \caption{
    \textcolor{black}{Quantitative (PSNR/SSIM) comparisons between VFI baselines and VFIformer's implementation with our strategy ($_{ours}$) on Vimeo90K~\cite{Alpher41} .
The best result is \first{boldfaced} . When combined with our approach, this method (VFIformer) can surpass current SOTA approaches. }}
    \resizebox{0.9\linewidth}{!}{\begin{tabular}{|c|c|c|c|c|c|}
    \hline
VFIformer  & VFIformer$_{ours}$  & EMA-VFI~\cite{zhang2023extracting}  &
DQBC~\cite{dqbc2023}  & IFRNet~\cite{ifrnet2022}  & EBME~\cite{ebme2022}  \\
\hline
\hline
36.38 & \first{36.69}
 & 36.64 & 36.57
 & 36.20 & 36.19 \\
 \hline
 0.9811 & \first{0.9826} & 0.9819 & 0.9817 & 0.9808 & 0.9810   \\
 \hline
 ABME~\cite{abme2021}  & SoftSplat~\cite{softsplat2020}  &SGM-VFI~\cite{liu2024sparse}  &TTVFI~\cite{liu2023ttvfi}  & InterpAny~\cite{zhong2024clearer}  & PerVFI~\cite{asymmetricble}  \\
 \hline
36.18 &	36.10 & 32.50 &36.54 & 35.51 & 33.60 \\
\hline
 0.9805 & 0.9700 &0.9689 &0.9819 & 0.9779 & 0.9256 \\
 \hline
\end{tabular}}
\label{tab:tab1-1}
\vspace{-0.1in}
\end{table*}

\subsection{Implementation Results}

We evaluate our proposed HRFFM with RDPs to enhance the performance of current representative VFI baselines, including VFIformer~\cite{Alpher19}, UPR-Net~\cite{Alpher39}, M2M-PWC~\cite{Alpher40}, \textcolor{black}{AMT~\cite{li2023amt},  Bim-VFI~\cite{seo2025bim}} and \textcolor{black}{PerVFI}~\cite{asymmetricble}. To ensure a fair comparison, we report results by implementing the officially released source code and training models under unified conditions on the same machine, rather than replicating results from the original papers. We maintain the original model architecture and loss function, incorporating our method into the feature encoder, as illustrated in Fig.~\ref{fig:framework1}.

\subsection{Comparison with VFI Baselines}

\noindent\textbf{Quantitative comparison.} Comparison results are presented in Tab.~\ref{tab:tab1}, where we integrate our proposed approach with VFI baselines to assess performance improvements. It is observed that almost all baselines exhibit enhanced results across all testing sets when our strategy is applied, with only a minimal increase in parameters and computation costs. Notably, our method demonstrates a substantial improvement of 0.31dB on Vimeo90K for the robust baseline, VFIformer (\textcolor{black}{presented in Tab.~\ref{tab:tab1-1}}). 
For other methods, there is an improvement of more than 0.1dB \textcolor{black}{on PSNR}, which is significant in the context of VFI tasks where performance has almost approached the upper limit. \textcolor{black}{Our approach achieves consistently lower FID scores, reflecting a high degree of statistical similarity to the target distribution. This is further corroborated by the LPIPS results, which confirm that our method produces superior image quality from a human-centric perceptual standpoint.}

Moreover, we conduct a comparison with several current SOTA VFI methods, including EMA-VFI~\cite{zhang2023extracting}, DQBC~\cite{dqbc2023}, IFRNet~\cite{ifrnet2022}, EBME~\cite{ebme2022}, ABME~\cite{abme2021}, SoftSplat~\cite{softsplat2020}, \textcolor{black}{SGM-VFI~\cite{liu2024sparse}, TTVFI~\cite{liu2023ttvfi},  InterpAny~\cite{zhong2024clearer}}, and \textcolor{black}{PerVFI}~\cite{asymmetricble} as outlined in Tab.~\ref{tab:tab1-1}. 
The results reveal that when integrated with our strategy, the chosen VFI baseline, i.e., VFIformer, can outperform these competitive SOTA approaches.

\textcolor{black}{
Furthermore, we conduct experiments on the large-scale benchmark LAVIB~\cite{stergiou2024lavib}. Specifically, we use LAVIB to evaluate our method alongside several representative VFI approaches. 
In addition to standard metrics, we also include FVD~\cite{unterthiner2019fvd}, which measures the distance between generated outputs and ground truth across multiple dimensions.
In particular, we randomly sample a sufficient number of training and evaluation examples from the corresponding datasets, ensuring that the train-test split follows the official setting and remains consistent across all competing methods.
As shown in Tab.~\ref{tab4}, our method continues to outperform others on this dataset.}

\begin{table}[t]
    \centering
    \caption{\textcolor{black}{The comparison between VFI baselines and their counterparts enhanced with our method, evaluated across more datasets and additional metrics.}}
    \label{tab4}
    \resizebox{1\linewidth}{!}{\begin{tabular}{|l|p{2.2cm}<{\centering}|p{2.2cm}<{\centering}|p{2.2cm}<{\centering}|}
    \hline
\multirow{2}{*}{Methods} & \multicolumn{1}{c|}{Vimeo90K} &
\multicolumn{1}{c|}{SNU-FILM$_{extreme}$} & \multicolumn{1}{c|}{LAVIB}   \\
\cline{2-4}
&FVD$\downarrow$   &FVD$\downarrow$ & PSNR/SSIM$\uparrow$ \\
\hline
\hline
UPR-Net	&  99.9108 &	179.31 
 &		31.86/0.9420   \\
\hline
\textbf{UPR-Net$_{ours}$}	&  \textbf{97.3529} &	 \textbf{176.85}
 &		\textbf{31.97/0.9563}   \\
\hline
\end{tabular}}
\label{tab:tab4}
\vspace{-0.2in}
\end{table}

\noindent\textbf{Qualitative comparison.} We present a visual comparison between the baselines and their counterparts combined with our approach, illustrated in Fig.~\ref{fig:rescompare}. Evidently, our strategy yields perceptual improvements by reducing undesirable artifacts and enhancing the accuracy of details.

\textcolor{black}{Moreover, even in the presence of occlusions (as shown in the foot region of Fig.~\ref{fig:tsne}), our approach can effectively distinguish the corresponding regions via RDPs obtained from SAM2, preserving clear boundaries and reducing interpolation errors in those areas. }

\begin{figure}[!t]
	\begin{center} 
		\includegraphics[width=0.98\linewidth]{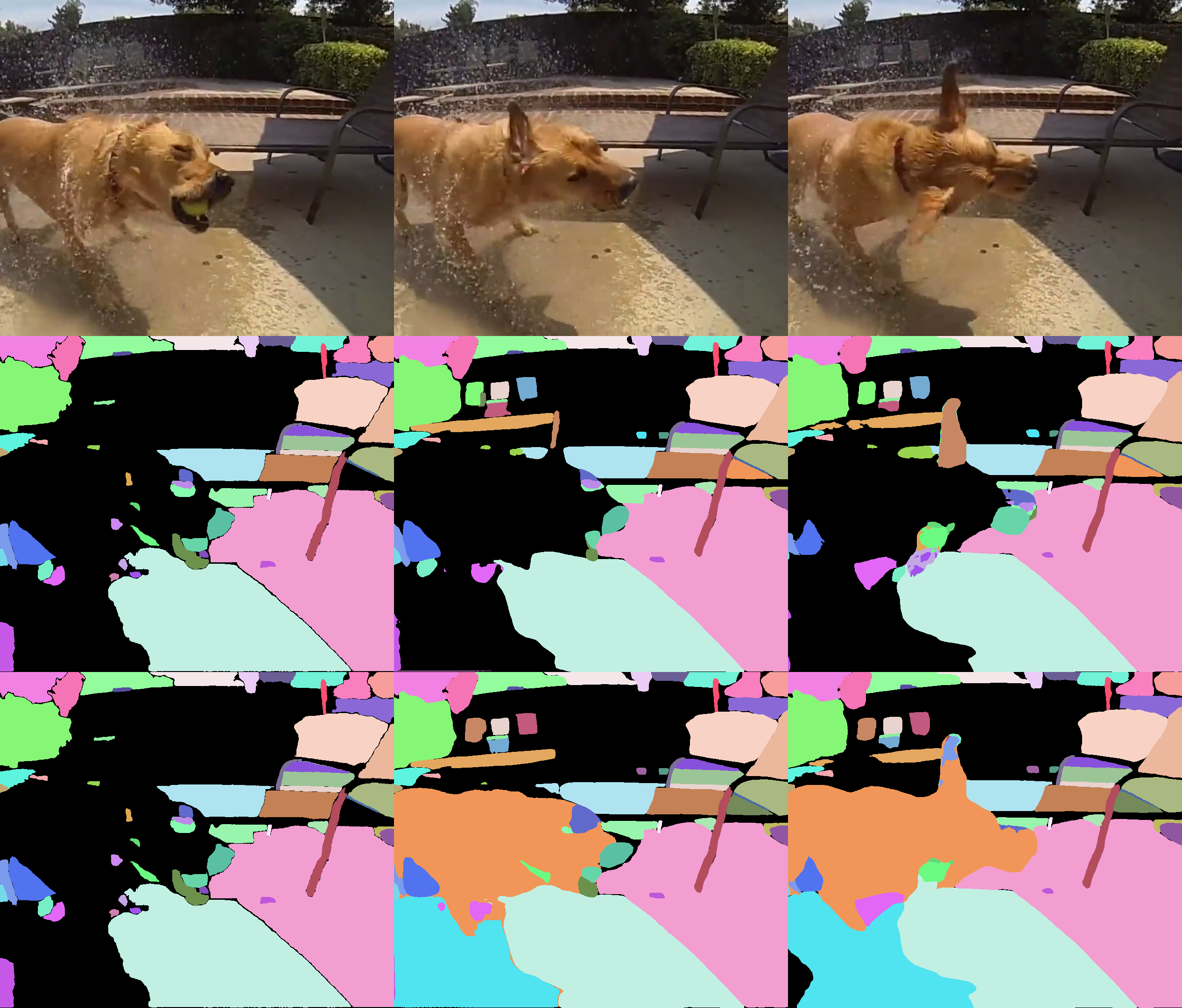}
	\end{center}
	\vspace{-0.05in}
	\caption{
	 Comparisons between baselines and ours in terms of the downstream video segmentation task. The top row is three continuous frames for visualization. The second and third rows are the segmentation results on the input and synthesized intermediate frames with the baseline and ours, respectively.
}
	\label{videosam}
	\vspace{-0.2in}
\end{figure}

\noindent\textbf{Evaluation with downstream tasks.} 
The VFI capability can be leveraged for various downstream tasks, including video segmentation. Large temporal gaps in videos can disrupt the effective propagation of semantic information. 
To assess the effect of our framework in terms of its impact on downstream video segmentation tasks, we employ the SOTA video segmentation approach SAM-Track\cite{cheng2023segment}.
The results, presented in Fig.~\ref{videosam}, showcase three consecutive frames in the first row, with segmentation results of synthesized intermediate frames generated by VFIformer and VFIformer$_{ours}$ in the second and third row, respectively. 
It is evident that the intermediate frames produced by our model exhibit more accurate segmentation. Our method's results enhance better temporal propagation among frames and can even rectify incorrect segmentation results in the first frame. For instance, the dog in the second row is not clearly separated from the shadow on the ground, whereas in the third row, the separation is more distinct.

\begin{figure}[t]
	\centering	
    \begin{subfigure}{1\linewidth}
    \centering
    \includegraphics[width=1.0\linewidth]{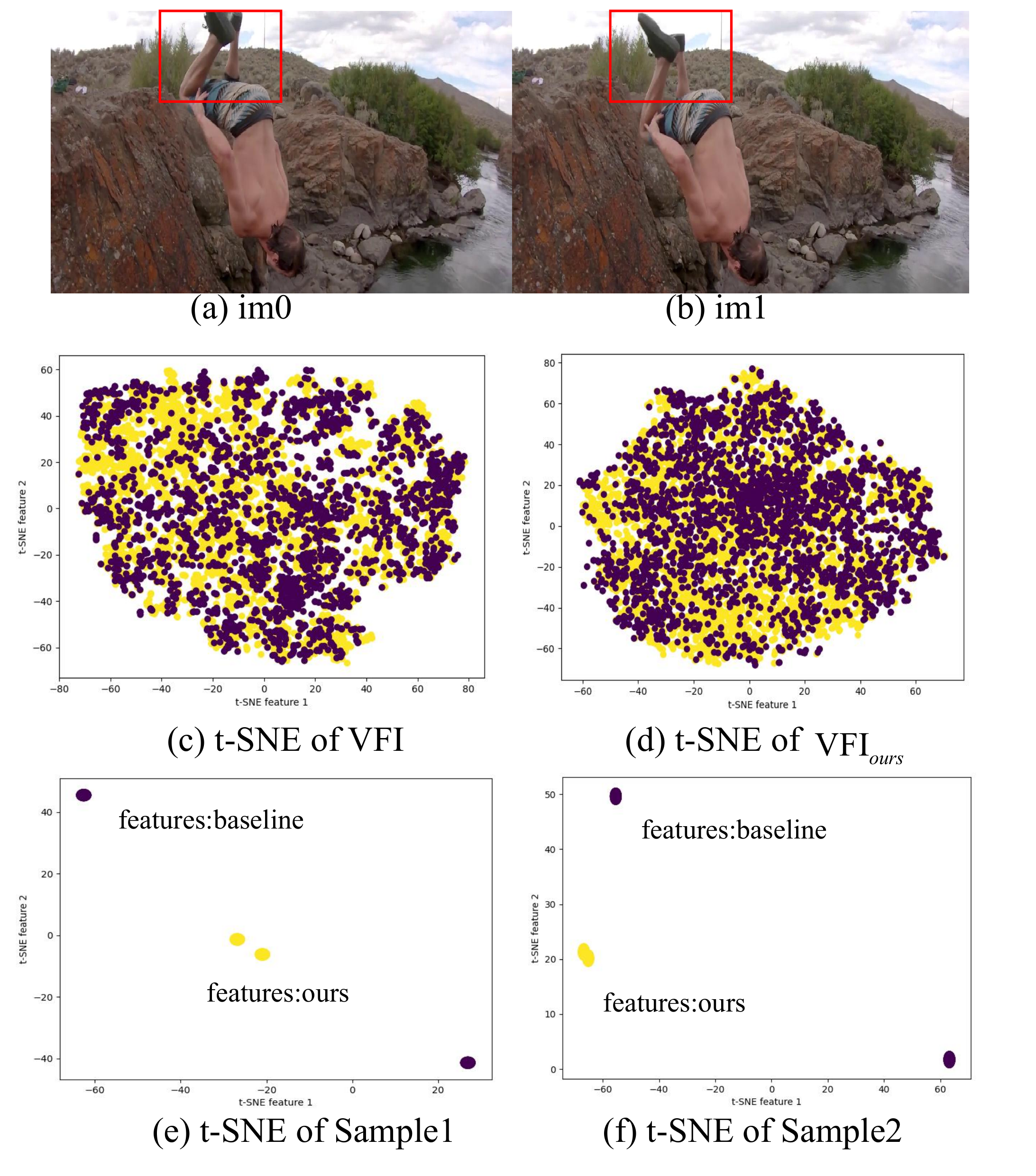}
    \end{subfigure}
    \hfill
    \vspace{-0.1in}
		\caption{\textcolor{black}{The t-SNE visualization comparison about the intermediate features between the baseline VFI and our enhanced version, VFI$_{ours}$. Two colors represent the feature distributions of the same object in two different frames. As shown, the left t-SNE figure (c) exhibits more isolated regions, indicating that the two feature distributions are more separated. In contrast, the right t-SNE figure  (d) shows a greater overlap between the distributions, demonstrating improved feature consistency across frames. 
        Moreover, as shown in the bottommost figure (e) and (f), we select two pixels within the red bounding box for feature comparison. The two dark-colored points represent the features of this pixel in the two consecutive frames synthesized by the VFI model, whereas the yellow and green points correspond to those generated by our method. Clearly, the feature distance between the two frames produced by our approach is substantially smaller.}
        }
		\label{fig:tsne}
\end{figure}

\noindent\textbf{Feature-level visualization for alignment analysis.} 
\textcolor{black}{
We propose a feature-level visualization strategy to illustrate the effectiveness of our method in achieving motion alignment and thus enhancing interpolation quality. To create this visualization, we first extract feature maps from the start and end frames and identify feature regions belonging to the same object, referred to as ``feature region A" and ``feature region B". We then apply the t-SNE technique to project all feature vectors from regions A and B into a 2D space. If alignment is successful, these feature vectors—representing the same object—should be closely aligned and overlapped. As shown in Fig.~\ref{fig:tsne}, feature vectors extracted by our encoder from regions A and B (the leg area within the red box in im0 and im1) exhibit stronger alignment compared to the baseline. }
\textcolor{black}{For clearer analysis, we select the feature of a single pixel within the red bounding box for comparison. As shown in the visualization, the distance between these two features produced by our method is significantly smaller, indicating that our approach effectively enhances feature alignment.
} \textcolor{black}{ Additionally, we compute the overlap mIoU between these two groups of feature vectors in the 2D space, with values of 32\% (ours) and 15\% (baseline), further confirming that our method achieves better feature alignment for the same object across different frames.}

\textcolor{black}{Moreover, we visualize the optical flow obtained from the deep features with a few layers. As shown in Fig.~\ref{fig:teaser}, different regions exhibit distinct flow values with clear boundaries between them. This indicates that the regions have different feature representations, as reflected by the variations in their decoded flow values.
This separation could facilitate interpolation by preventing incorrect alignment across varying objects.
}

\noindent\textbf{Comparison with generative-model-based methods.} 
\textcolor{black}{As indicated in the introduction, our method differs in its objectives from generative-model-based approaches, which often fall short in achieving high fidelity. To support this claim, we provide the evaluation results of a SOTA and representative generative frame interpolation method, FCVG~\cite{zhu2025generative}. Its performance on various datasets is as follows: 25.52/0.8554 on Vimeo90K, 18.35/0.6876 on UCF101, 24.29/0.8021 on SNU-FILM (Easy), 21.85/0.7516 on SNU-FILM (Medium), 19.89/0.7237 on SNU-FILM (Hard), and 18.44/0.6618 on SNU-FILM (Extreme).
These results are clearly lower than those of restoration-based methods, as shown in Tab.~\ref{tab:tab1}. Thus, while generative methods are capable of producing photo-realistic frames, their fidelity to the ground truth remains relatively low.}

\subsection{Ablation Study}
\label{sec:absd}
In this section, we perform various ablation studies to examine different components in our proposed method.
The first ablation test (\textcolor{black}{ the corresponding results in Tab.~\ref{table:sil2h}}) was trained to completion, while all subsequent ablation experiments were truncated at 100,000 iterations.

\noindent{\textbf{Effect of HRFFM}.}
We have designed a comprehensive framework that leverages semantic segmentation features derived from \textcolor{black}{SAM2}, specifically RDPs, to distinguish objects in images. Additionally, we meticulously crafted a module that fuses image and segmentation features to enhance the accuracy of flow estimation, thereby improving the performance of intermediate frame interpolation.
In this section, we explore two crucial questions: (1) whether RDPs are indispensable components of the proposed framework, and (2) whether the performance gains of HRFFM are solely due to the increased network depth or can be attributed to the specific design of the framework. To address these questions, we conducted two experiments.
First, we removed the RDPs from the framework, as illustrated in $s'_{i,l}$ of Fig.~\ref{fig:fig3}. Second, we replaced our HRFFM with a simple two-layer convolutional neural network while keeping all other conditions unchanged. The results of these experiments are presented in Tab.~\ref{table:sil2h}, where UPR-Net$_{NoRDPs}$ and UPR-Net$_{2conv}$ correspond to these two experimental settings, respectively.
The results clearly show that the model incorporating segmentation features outperforms the model relying solely on image features, as demonstrated by the comparative performance of UPR-Net$_{ours}$ and UPR-Net$_{NoRDPs}$. Furthermore, our proposed framework exhibits a substantial performance advantage over the simplistic two-layer convolutional neural network, indicating that the observed improvement is not merely a consequence of increased network depth, but rather the result of the synergistic integration of image and segmentation features.

\begin{table}[t]
\centering
{\caption{Ablation study results. We calculate PSNR, LPIPS, and NIQE values of different ablation settings in the Vimeo90K dataset. \first{boldfaced} indicates the best results.
}
		\label{tab:ablation}}
		\resizebox{0.7\linewidth}{!}{
\begin{tabular}{|p{2cm}<{\centering}|c|c|c|}
    \hline
    Settings     & PSNR     & LPIPS & NIQE \\
    \hline
    \hline
    Ours with O.H. & 35.52  & 0.023
	& 26.95\\
Ours with L.E. & 35.40  & 0.0233
	& 27.15 \\
Ours w/o S.O. & 35.52  & 0.0152
 &	24.09 \\
Ours w/o R.L. & 35.54  & 0.0154
 &	24.04 \\
Ours with CNN & 35.39  & 0.0153
 &	24.19 	\\
Ours with Trans. & 35.53  & 0.0154
 &	24.00 \\
Full & \first{35.57}  & \first{0.0148}
&	\first{23.88} \\
    \hline
\end{tabular}}
\end{table}	

\begin{table}[t]
\centering
\caption{Ablation study results for the proposed strategy. We calculate LPIPS values of different ablation settings in the challenging high-resolution SNU-FILM dataset. 
		}
	\label{tab:ablation2}
	\resizebox{1.0\linewidth}{!}{
\begin{tabular}{|p{3cm}<{\centering}|c|c|c|c|}
\hline
Settings     & Easy
     & Medium
 & Hard & Extreme\\
    \hline \hline
    Ours with O.H. & 0.0191	&0.0337 &	0.0633 &	0.1178
\\
Ours with L.E. & 0.0195 &	0.034 &	0.0637 &	0.1159
 \\
Ours w/o S.O. &0.012 &	0.0218	& 0.0444 &	0.0898
\\
Ours w/o R.L. & 0.0123 &	0.0222 &	0.0452 &	0.0911
 \\
Ours with CNN &0.012 &	0.0218 &	0.0446 &	0.0901
	\\
Ours with Trans. & 0.0117 &	0.0216 &	0.0447 &	0.0906
\\
Full & \first{0.0111} &	\first{0.0205}	& \first{0.0426} &	\first{0.0877}
 \\
  \hline
\end{tabular}}
	
\end{table}

\begin{table*}[t]
    \setlength{\belowcaptionskip}{0pt}
    \renewcommand\arraystretch{1.15}
    \centering
     \caption{Quantitative (PSNR/SSIM) comparisons between UPR-Net , UPR-Net$_{ours}$ ,UPR-Net$_{NoRDPs}$ and UPR-Net$_{2conv}$ on Vimeo90K, UCF101, and SNU-FILM.}
   	\resizebox{1.0\linewidth}{!}{
    \begin{tabular}{|p{3cm}<{\centering}|p{1.8cm}<{\centering}|p{1.8cm}<{\centering}|p{1.8cm}<{\centering}p{1.8cm}<{\centering}p{1.8cm}<{\centering}p{1.8cm}<{\centering}|}
    \hline
   \multirow{2}{*}{Methods} & \multirow{2}{*}{Vimeo90K} &
\multirow{2}{*}{UCF101} & \multicolumn{4}{c|}{SNU-FILM} \\
\cline{4-7}
& &  &  Easy & Medium & Hard & Extreme\\
    \hline
    \hline
    UPR-Net~\cite{Alpher39}	&36.02/0.9800&	35.40/0.9698&		40.40/0.9910&	36.15/0.9797&	30.70/0.9364&	25.53/0.8631 \\
\textbf{UPR-Net$_{ours}$}	&\textbf{36.19/0.9806}&	\textbf{35.45/0.9699}&		\textbf{40.43/0.9911}&	\textbf{36.19/0.9798}&	\textbf{30.80/0.9370}&	\textbf{25.64/0.8643} 
\\
UPR-Net$_{NoRDPs}$ &36.18/0.9806&	35.44/0.9699&		40.37/0.9910&	36.15/0.9797&	30.73/0.9370&	25.57/0.8640 
\\
UPR-Net$_{2conv}$ &  36.08/0.9803&	35.41/0.9699&		34.02/0.9745&	32.81/0.9661&	29.31/0.9279&	24.99/0.8564 
\\
\hline
\end{tabular}}
\label{table:sil2h}
\end{table*}

\noindent\textbf{Effect of Mixture Gaussian embedding.} 
Mixture Gaussian embedding serves as a crucial representation for distinguishing objects between two frames, playing a pivotal role in adapting \textcolor{black}{SAM2} outputs for an arbitrary number of instances. To investigate the impact of Mixture Gaussian embedding, we replaced it with alternative methods, including naive one-hot encoding or learnable embeddings. Both alternatives require assuming a maximum instance number, denoted as ``Ours with O.H.'' and ``Ours with L.E.'', respectively. The results, presented in Tab.~\ref{tab:ablation}, indicate that their performance is lower than the results achieved with Mixture Gaussian embedding, highlighting the effect of the proposed approach outlined in Sec.~\ref{sec:sec31}.

\begin{figure*}[t]
    \begin{center}
    \includegraphics[width=1.0\linewidth]{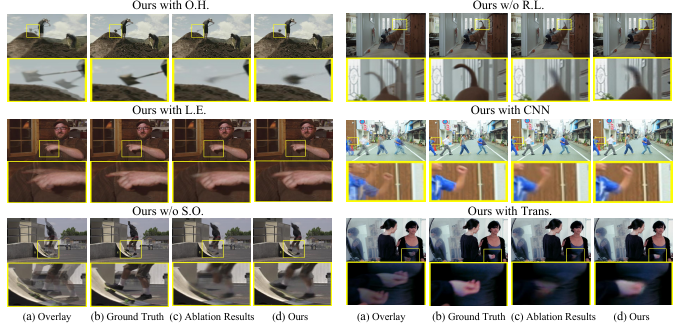}
	\end{center}
        \centering
	\caption{
Visual comparisons of ablation studies on Vimeo90K~\cite{Alpher41}. 
}	\label{fig:abalationpic}
     \end{figure*}

\noindent\textbf{Effect of softmax operation and residual learning in HRFFM.} 
After the feature extraction for RDP in each layer, the softmax operation ensures the consistency of feature representations at different scales. Additionally, to mitigate the impact of SAM2 errors on subsequent feature fusion, a residual learning component is incorporated after RDPFN. To assess their effectiveness, we trained two models without the softmax operation and residual learning, labeled as ``Ours w/o S.O.'' and ``Ours w/o R.L.'', respectively. As depicted in Tab.~\ref{tab:ablation}, the performance of both models is lower than the original full setting, underscoring the rationality of the softmax operation and residual learning in HRFFM.

\begin{table}[t]
    \setlength{\belowcaptionskip}{0pt}
    \renewcommand\arraystretch{1.15}
    \centering
     \caption{The ablation study results for the Gaussian mixture channel choice.
 Different Gaussian mixture channels' results on three datasets, Vimeo90K\cite{Alpher41}, UCF101~\cite{Alpher42}, and Middlebury\cite{middlebury}, are given.}
    
    \resizebox{1.0\columnwidth}{!}{
    \begin{tabular}{|p{2cm}<{\centering}|c|c|c|}
    \hline
    Gaussian mixture channels&Vimeo90K&UCF101&Middlebury\\
    \hline
    \hline
        3 & \textbf{35.57}/\textbf{0.9783}  & \textbf{35.39}/\textbf{0.9696}& \textbf{37.49}/\textbf{0.9854} \\
        \hline
        24 &35.45/0.9780  & 35.31/0.9695& 37.32/0.9847\\
        \hline
        96 &35.39/0.9777  & 35.32/0.9695& 37.28/0.9848 \\
    \hline
 \end{tabular}}
\label{gaussian}
\vspace{-0.1in}
\end{table}

\noindent\textbf{Effect of parallel CNN and transformer blocks in RDPFN.}
RDPFN is designed to leverage both long- and short-range dependencies, formulating normalization parameters for regions with varying shapes and areas. To demonstrate the effectiveness of this parallel setting, we trained two models with only a convolutional layer and a Transformer layer in RDPFN, labeled as ``Ours with CNN'' and ``Ours with Trans.'', respectively. The results in Tab.~\ref{tab:ablation} indicate that removing either component leads to an overall performance degradation, underscoring the necessity of the parallel CNN and Transformer strategy in formulating suitable region-aware normalization parameters. To further validate the conclusions from the ablation experiments, we conduct additional ablation experiments on the SNU-FILM datasets (shown in Tab.~\ref{tab:ablation2}). These results also demonstrate that our proposed strategy outperforms the ablated settings.
In addition to quantitative comparisons, we also present visual comparisons. As shown in Fig.~\ref{fig:abalationpic}, the intermediate frames generated by six ablation studies and our method are shown in the last two columns. Obviously, our method produces better results than the others.

\begin{table*}[t]
    \setlength{\belowcaptionskip}{0pt}
    \renewcommand\arraystretch{1.15}
    \centering
     \caption{Quantitative (PSNR/SSIM) comparisons between UPR-Net and UPR-Net$_{ours}$ on Vimeo90K, UCF101, and SNU-FILM, whose data is degraded by the low-light. }

   \resizebox{1.0\linewidth}{!}{
    \begin{tabular}{|p{2.2cm}<{\centering}|p{1.8cm}<{\centering}|p{1.8cm}<{\centering}|p{1.8cm}<{\centering}p{1.8cm}<{\centering}p{1.8cm}<{\centering}p{1.8cm}<{\centering}|}
    \hline
   \multirow{2}{*}{Methods} & \multirow{2}{*}{Vimeo90K} &
\multirow{2}{*}{UCF101} & \multicolumn{4}{c|}{SNU-FILM} \\
\cline{4-7}
& &  &  Easy & Medium & Hard & Extreme\\
    \hline
     \hline
    UPR-Net& 42.06/0.9870 & 40.15/0.9759 & 44.49/0.9937 & 40.35/0.9850& 34.97/0.9503 &29.81/0.8895
\\
    UPR-Net$_{ours}$ &  \textbf{42.19}/\textbf{0.9877}	&\textbf{40.38}/\textbf{0.9763}	&	\textbf{44.64}/\textbf{0.9938}&	\textbf{40.51}/\textbf{0.9852}&	\textbf{35.17}/\textbf{0.9512}	&\textbf{30.07}/\textbf{0.8910}
\\
    \hline
\end{tabular}}
\label{table:lowlight}
\end{table*}

\noindent{\textbf{The influence of Gaussian mixture channels.}
As described in Sec.~\ref{sec:defr}, the RDPs are represented as 3D Gaussian mixtures. We conducted an ablation study to explore the impact of different Gaussian mixture dimensions. The experiments were carried out using the UPR-Net framework, with each ablation setting trained for 100,000 iterations. The results, presented in Tab.~\ref{gaussian}, indicate that across all three datasets, when the Gaussian mixture dimension is set to 24 or 96, the interpolation performance is not as effective as with our original setting of 3. This confirms that increasing the Gaussian mixture dimension does not necessarily yield better results. We think there are two main reasons for this phenomenon. First, increasing the dimension raises model complexity, resulting in more parameters to optimize, which is challenging given the fixed number of training samples across experiments. Second, higher dimensionality introduces more randomness in the RDPs, making the learning process harder due to an increased degree of ill-posedness. In summary, increasing the dimension intensifies the learning difficulty, causing poor alignment and consequently incorrect motion interpolation results.

\begin{table}[t]
    \centering
    \caption{ Comparison of our method using SAM2 and alternative segmentation models, i.e.,
    including the original SAM2 (UPR-Net$_{ours}$), a semantic segmentation variant (UPR-Net$_{semantic}$), and a lightweight version, Tiny-SAM (UPR-Net$_{tinysam}$).}
\large
    \label{tab3}
    \resizebox{\linewidth}{!}{\begin{tabular}{|p{4cm}<{\centering}|p{3.5cm}<{\centering}|p{3.5cm}<{\centering}|}
    \hline
\multirow{2}{*}{Methods}& \multirow{1}{*}{Vimeo90K} &
\multirow{1}{*}{UCF101} \\
\cline{2-3} 
& PSNR$\uparrow$ & PSNR$\uparrow$ \\
\hline
\hline
UPR-Net~\cite{Alpher39}	& 35.49  & 35.27 \\
\textbf{UPR-Net$_{ours}$}	&\textbf{35.78
}&	\textbf{35.39}	 \\
\textbf{UPR-Net$_{semantic}$}	& 35.65 &	35.30 \\
\textbf{UPR-Net$_{tinysam}$}	& 35.77 &	35.39
   \\
\hline
\end{tabular}}
\label{tab:tab3}
\end{table}

\noindent{\textbf{The results with efficient SAM2.}
\textcolor{black}{The primary goal of this paper is to demonstrate the effectiveness of using RDPs from SAM2 to enhance the performance of existing VFI methods—an approach that has not been explored or implemented in prior work. Meanwhile, we also note that inference efficiency is crucial for real-world applications. Notably, recent advancements in SAM, such as Tiny-SAM~\cite{shu2025tinysam}, have significantly improved inference speed. Therefore, we have conducted an additional experiment using Tiny-SAM to obtain our RDPs. The results, presented in Tab.~\ref{tab:tab3}, show that our method with Tiny-SAM still improves baseline performance, highlighting its potential for practical use. 
Note that due to limited time and resources, all experiments in  Tab.~\ref{tab:tab3} were truncated at 200,000 iterations. \textcolor{black}{
To validate our claims, we evaluated the inference time of SAM 2 and TinySAM at a standard resolution of 
$1024\times 1024$ using a mid-to-high-end GPU (NVIDIA A100), as shown in Tab.~\ref{tab:efficiency_comparison}.}
}

\begin{table}[h]
\centering
\caption{Comparison of Computational Efficiency: SAM 2 vs. TinySAM}
\label{tab:efficiency_comparison}
\resizebox{0.50\textwidth}{!}{
\begin{tabular}{@{}lccc@{}}
\toprule
\textbf{Model Variant} & \textbf{Parameters (M)} & \textbf{Inference Time (ms)} & \textbf{Throughput (FPS)} \\ \midrule
SAM 2 (H-ViT)          & 224.4                   & 25.8                        & 38.7                     \\
SAM 2 (L-ViT)          & 107.4                   & 15.6                        & 64.1                     \\
\textbf{TinySAM}       & \textbf{5.7}            & \textbf{2.1}                & \textbf{476.2}           \\ \bottomrule
\end{tabular}}
\end{table}

\noindent{\textbf{The results with semantic segmentation priors.}
\textcolor{black}{We think that semantic information alone may not be effective for interpolation, as objects sharing the same semantic label can exhibit different motions. Therefore, instance-level segmentation is more suitable in this context, and we choose SAM2. Here, we conduct an experiment in which semantic labels are used to replace SAM2 outputs for generating RDPs. 
In particular, we utilize PSPNet~\cite{zhao2017pyramid} as a representative semantic segmentation model, trained on ADE20K~\cite{zhou2017scene} with its diverse and extensive category set.
As shown in Tab.~\ref{tab3}, this approach results in lower performance compared to our original implementation, supporting our claim.}

\subsection{Comparison in Low-light Conditions with VFI Baselines}

Additionally, we compared our method against the original baselines in practical yet challenging scenarios, such as low-light environments. Tab.~\ref{table:lowlight} presents the quantitative comparison results, while Fig.~\ref{fig:low} provides visual comparisons. For these evaluations, we applied low-light degradations to the input frames from three datasets before testing. The comprehensive comparisons confirm the effectiveness of our framework in improving interpolation performance under low-light conditions.

This experiment also demonstrates that our method is robust to lighting variations (a common challenge in video interpolation) largely due to our use of SAM2 outputs, which are inherently resilient to lighting changes.

\begin{figure}[!t]
	\begin{center} 
\includegraphics[width=1.0\linewidth]{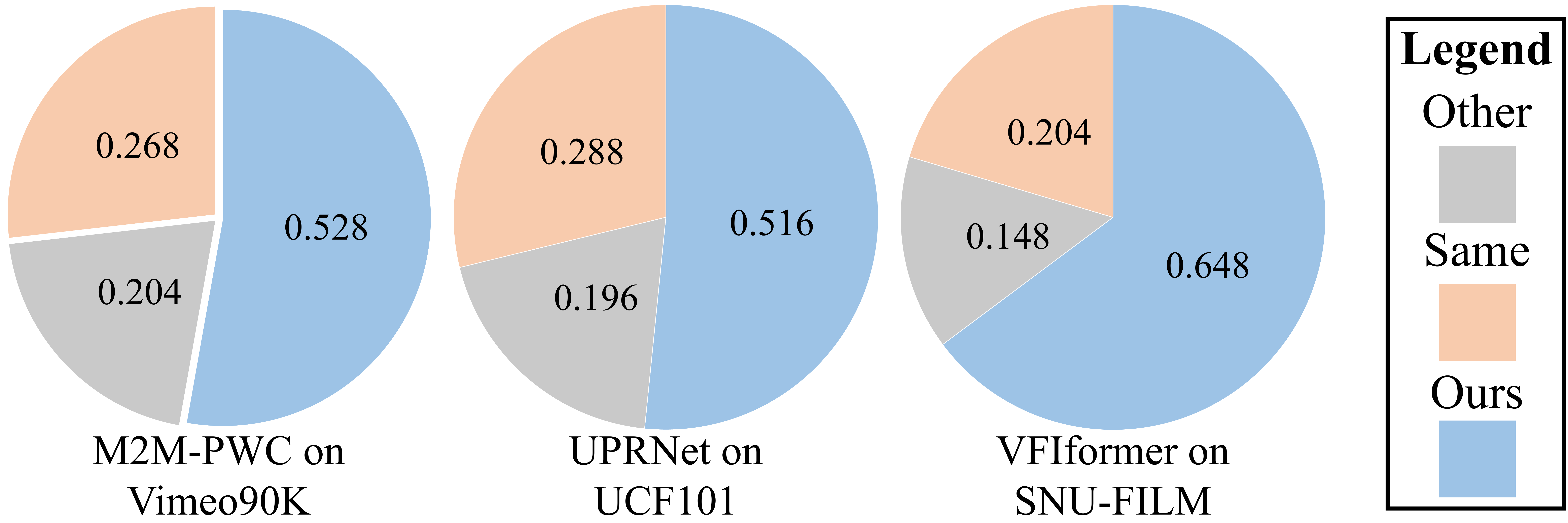}
	\end{center}
	\vspace{-0.1in}
        \centering
	\caption{
		The results of the user study, which summarize that the results enhanced with our strategy, are preferred by participants compared with the baselines' results.
	}
	\label{us_tbl}
\end{figure}

\begin{figure}[t]
    \begin{center}
	\includegraphics[width=1.0\linewidth]{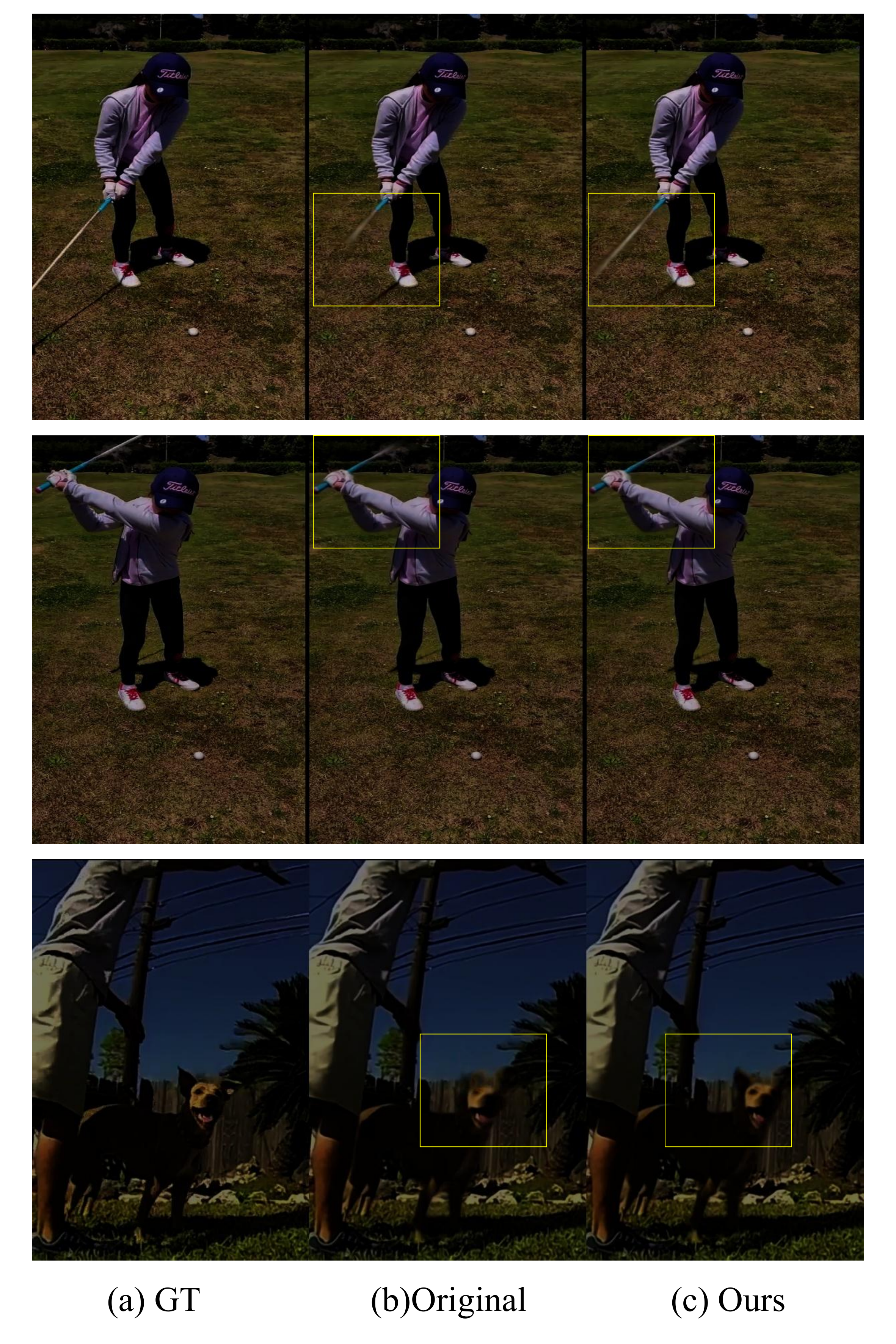}
	\end{center}
	\vspace{-0.1in}
	\caption{Visual comparison of UPR-Net~\cite{Alpher39} and UPR-Net$_{ours}$ on SNU-FILM~\cite{Alpher43} whose data is degraded by the low-light.
 }	
 \label{fig:low}
 \end{figure}

\subsection{User Study}

To assess the effectiveness of our proposed framework through subjective evaluation, we carried out an extensive user study involving 50 participants via online questionnaires.

To execute the user study, we randomly gathered 20 videos for each testing set and employed the AB-test methodology. Participants were presented with an example for assessment, featuring input two frames, baseline results, and our results. Their task was to choose the superior one based on the consistency between the interpolated results and input frames, taking into account details and artifacts in the interpolated frame. The positions of our results and baseline results were randomized during each evaluation. Each participant compared 5 pairs for a specific method on a given dataset, with the options to indicate whether ours was better, the baseline was better, or if they were the same (without knowledge of which method was ours). Each participant completed 15 tasks (3 methods $\times$ 5 videos), and on average, it took approximately 15 minutes for a participant to finish the user study.

Fig.~\ref{us_tbl} displays the results of the user study, revealing that our method received more selections from participants compared to all the baselines. While some participants opted for the ``same'' option, this is primarily attributed to the resolution of the testing images. Higher resolution tends to amplify differences, as observed in the results from the SNU-FILM dataset. This underscores that our method can enhance the human subjective perception of baselines.

\textcolor{black}{In addition, We would like to emphasize the principles of our experiment:
\begin{itemize}
\item Strict Fairness: All video samples were presented at the same resolution to eliminate display bias.
\item Randomized Order: Both the sequence of pictures and the positioning of methods (Ours vs. Competitors) were fully randomized to prevent order effects.
\end{itemize}
}

\section{Limitations}

While our proposed method has achieved commendable performance improvement on multiple datasets, there are several limitations that we aim to address in future work. First, we plan to investigate more lightweight approaches, such as employing advanced networks to further reduce the parameter and computation cost.

\section{Conclusion}
In this work, we introduced a plug-and-play module designed to enhance the performance of existing VFI approaches. We innovatively designed RDPs using \textcolor{black}{SAM2} and implemented the HRFFM to integrate them into VFI methods. Extensive experiments demonstrate that our strategy significantly improves the performance of current VFI methods, achieving SOTA results across multiple well-recognized benchmarks.

\section{Acknowledgement}
This research is supported by the Key Research Project of Zhejiang Lab (Grant No. 2025SSYS0004), the Key R\&D Program of Zhejiang (Grant No. 2025SSYS0001) and  the Zhejiang Leading Innovative and Entrepreneurial Team (Chief Scientist) under Grant No. 2025R01002.

\bibliographystyle{splncs04}
\bibliography{main}


\begin{IEEEbiography}[{\includegraphics[width=1in,height=1.25in,clip]{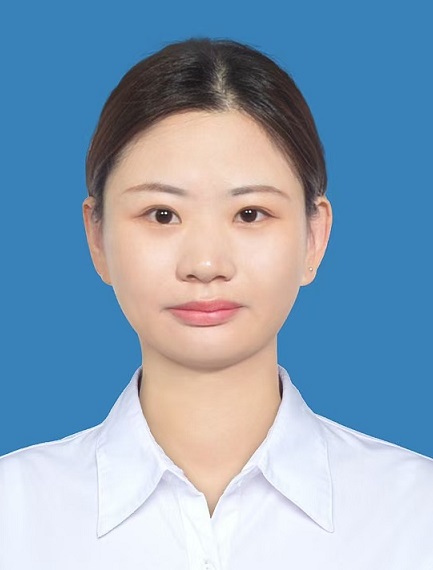}}]{Yan Han} received her bachelor degree from Huazhong Agricultural University, and received master degree from Wuhan University in 2019. She is currently working at Zhejiang Lab, where her research focuses on computer vision, image processing, and multimodal models,etc.
\end{IEEEbiography}

\begin{IEEEbiography}[{\includegraphics[width=1in,height=1.25in,clip]{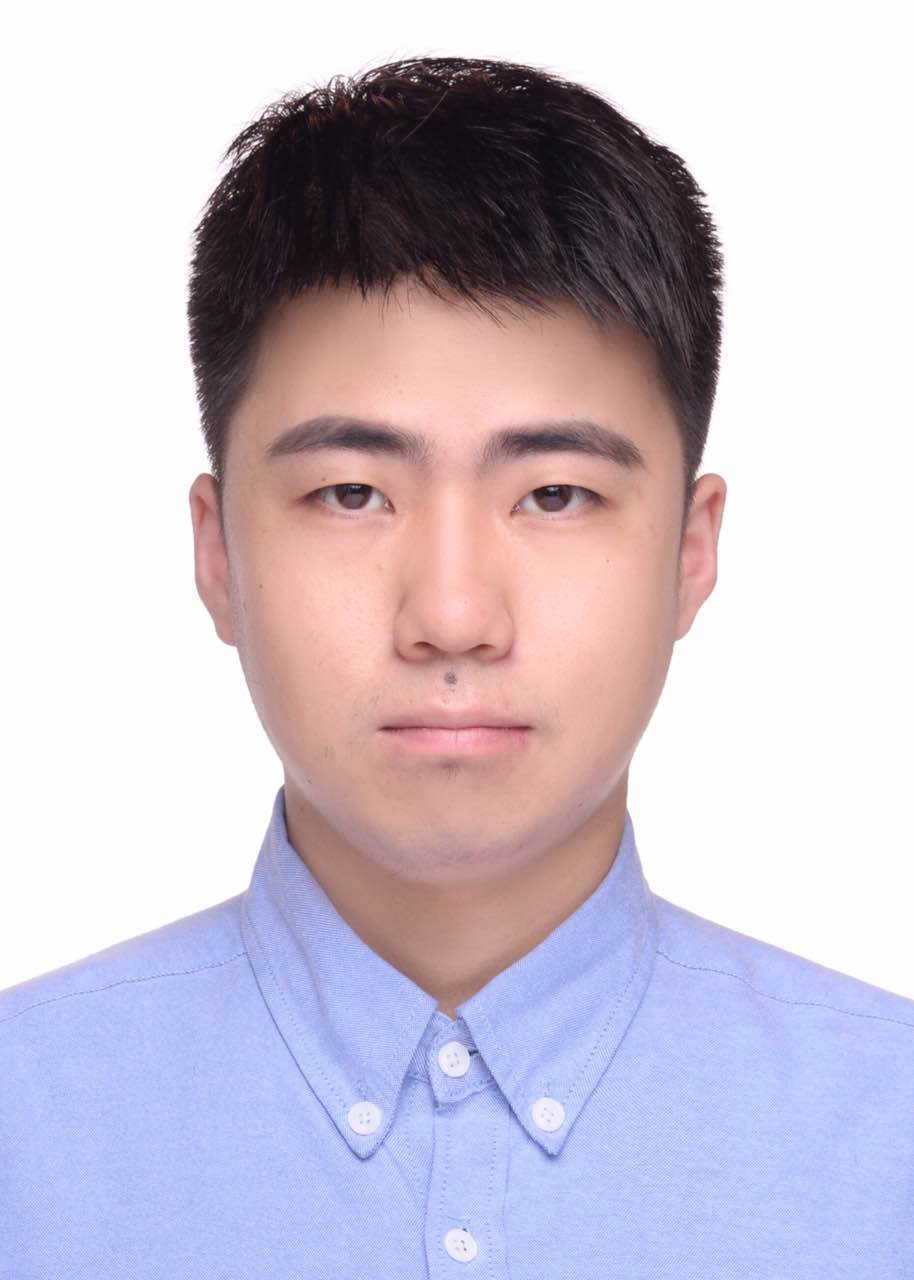}}]{Xiaogang Xu} is a postdoc research fellow in the Chinese University of Hong Kong. He received his Ph.D. degree from CUHK in 2022 and bachelor degree from Zhejiang University in 2018. In 2023, he is a research scientist in Zhejiang Lab and meanwhile a ZJU100 Young Professor at ZJU. He obtained the Hong Kong PhD Fellowship in 2018. He serves as a reviewer for CVPR, ICCV, ECCV, Neurips, ICLR, TPAMI, TIP, IJCV, etc. His research interest includes deep learning, computational photography, AIGC, large models.

\end{IEEEbiography}

\begin{IEEEbiography}[{\includegraphics[width=1in,height=1.25in,clip]{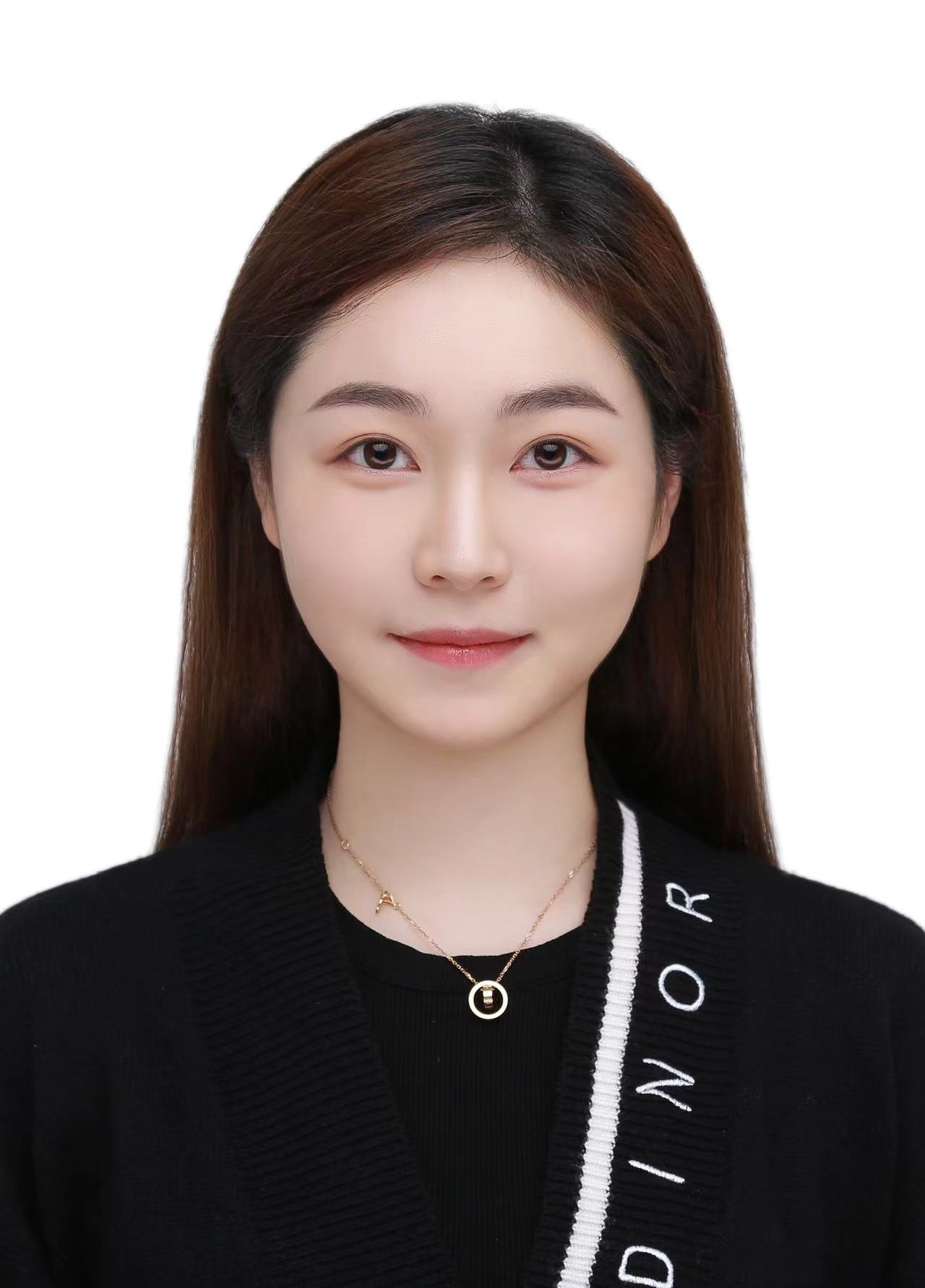}}]{Yingqi Lin} received her bachelor degree from Shih Hsin University, and received master degree from University of Southern California in 2021. She is currently working at Zhejiang Lab, where her research focuses on computer vision, image processing, and multimodal models,etc.
\end{IEEEbiography}

\begin{IEEEbiography}[{\includegraphics[width=1in,height=1.25in,clip]{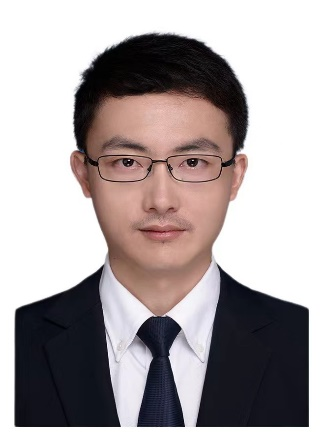}}]{Jiafei Wu} received the B.S. degree from JXUFE in 2010, the M.S. degree and Ph.D. degree from the University of Hong Kong in 2012 and 2017, respectively. He has been a senior engineer, manager and deputy director from 2018 to 2023 in SenseTime. He is currently with the Zhejiang Lab. His research interests include deep learning, trustworthy AI, embedded system, and computational intelligence.
\end{IEEEbiography}

\begin{IEEEbiography}[{\includegraphics[width=1in,height=1.25in,clip]{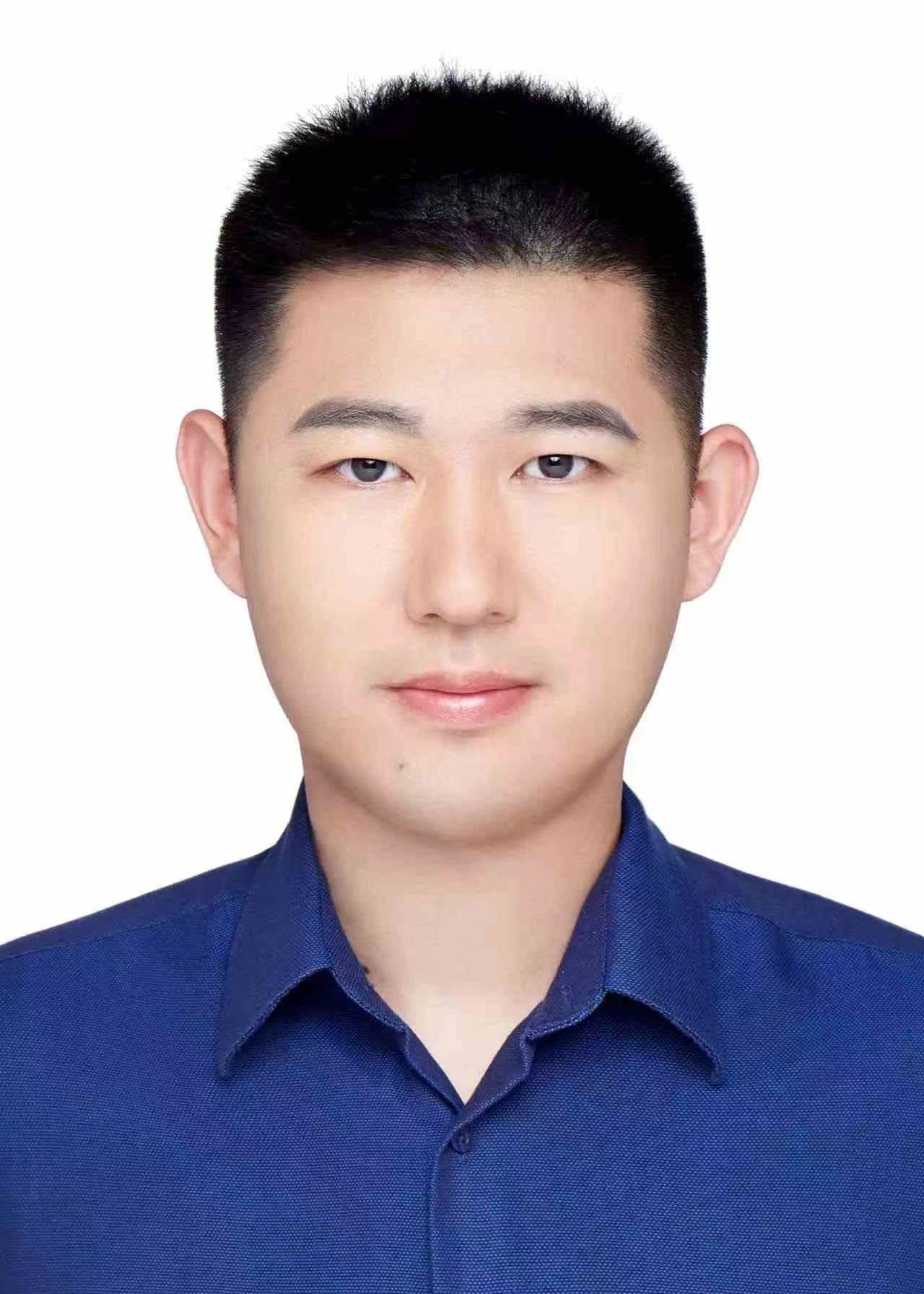}}]{Zhe Liu} received the BS and MS degrees from
Shandong University, China, in 2008 and 2011,
respectively, and the PhD degree from the Lab-
oratory of Algorithmics, Cryptology and Secu-
rity, University of Luxembourg, Luxembourg, in
2015. He is a professor with Zhejiang Lab. His
research interests include security, privacy and cryptography solutions for the Internet of Things.
\end{IEEEbiography}

\begin{IEEEbiography}[{\includegraphics[width=1in,height=1.25in,clip]{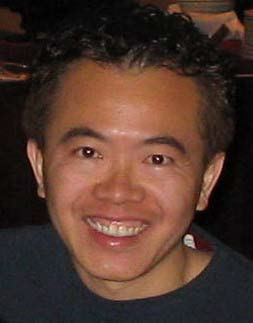}}]{Ming-Hsuan Yang} is a Professor of Electrical Engineering and Computer Science at the University of California, Merced. Yang serves as a program co-chair of the ICCV in 2019.  He received the Best Paper Award at ICML 2024, Longuet-Higgins Prize at CVPR 2023, Best Paper Honorable Mention at CVPR 2018, Nvidia Pioneer Research Award in 2018 and 2017, NSF CAREER award in 2012 and Google Faculty Award in 2009. He is a Fellow of the IEEE, ACM and AAAI.
\end{IEEEbiography}

\vfill

\end{document}